\documentclass[
]{ceurart}

\usepackage{listings}
\usepackage{latexsym}
\usepackage{amssymb}
\usepackage{amsmath}
\usepackage{amsthm}
\usepackage{booktabs}
\usepackage{enumitem}
\usepackage{graphicx}
\usepackage{color}

\usepackage{latexsym}
\usepackage{amssymb}
\usepackage{amsmath}
\usepackage{amsthm}
\usepackage{booktabs}
\usepackage{enumitem}
\usepackage{graphicx, subcaption}
\usepackage{algorithm}
\usepackage[noend]{algpseudocode}
\usepackage{varwidth} 

\algrenewcommand\algorithmicindent{0.8em}
\algrenewcommand\alglinenumber[1]{\footnotesize #1}
\usepackage{color}
\usepackage[table,xcdraw]{xcolor}

\usepackage{array}
\usepackage{multirow}
\usepackage{comment}
\usepackage{float}
\usepackage{makecell}
\usepackage{placeins}

\usepackage{xcolor}
\definecolor{myPurple}{RGB}{128, 0, 128}  
\newcommand{\new}[1]{\textcolor{black}{#1}}

\newtheorem{theorem}{Theorem}

\usepackage{amsmath}
\usepackage{amssymb}
\usepackage{mathtools}
\usepackage{amsthm}

\usepackage[capitalize,noabbrev]{cleveref}

\theoremstyle{plain}
\theoremstyle{definition}

\theoremstyle{remark}

\newcommand{\BibTeX}{B\kern-.05em{\sc i\kern-.025em b}\kern-.08em\TeX}

\begin{document}

\copyrightyear{2025}
\copyrightclause{Copyright for this paper by its authors.
  Use permitted under Creative Commons License Attribution 4.0
  International (CC BY 4.0).}

\conference{TRUST-AI: The European Workshop on Trustworthy AI. Organized as part of the European Conference of Artificial Intelligence - ECAI 2025. October 2025, Bologna, Italy.}

\title{Learning Fairer Representations with FairVIC}

\author[1]{Charmaine Barker}[%
orcid=0000-0002-2337-043X,
email=charmaine.barker@york.ac.uk,
]

\cormark[1]

\author[1]{Daniel Bethell}[%
orcid=0000-0002-0685-5312,
email=daniel.bethell@york.ac.uk,
]

\author[1]{Dimitar Kazakov}[%
orcid=0000-0002-0637-8106,
email=dimitar.kazakov@york.ac.uk,
]

\address[1]{Department of Computer Science, University of York, York, United Kingdom}

\cortext[1]{Corresponding author.}

\begin{abstract}
Mitigating bias in automated decision-making systems, particularly in deep learning models, is a critical challenge due to nuanced definitions of fairness, dataset-specific biases, and the inherent trade-off between fairness and accuracy.  To address these issues, we introduce FairVIC, an innovative approach that enhances fairness in neural networks by integrating variance, invariance, and covariance terms into the loss function during training. Unlike methods based on predefined criteria, FairVIC abstracts fairness to minimise dependency on protected attributes. We evaluate FairVIC against comparable bias mitigation techniques on benchmark datasets, considering both group and individual fairness, and conduct an ablation study on the accuracy-fairness trade-off. FairVIC demonstrates significant improvements ($\approx70\%$) in fairness across all tested metrics without significantly compromising accuracy ($\approx-5\%$), thus offering a robust, generalisable solution for fair deep learning across diverse tasks and datasets.
\end{abstract}


\maketitle

\section{Introduction}
\label{sec:Introduction}
With the ever-increasing utilisation of Artificial Intelligence (AI) in everyday applications, neural networks have emerged as pivotal tools for automated decision making systems in sectors such as healthcare~\citep{medicalADM}, finance~\citep{financeADM}, and recruitment~\citep{recruitmentADM}. However, bias in the data--stemming from historical inequalities, imbalanced distributions, or flawed feature representations--are often learned by these models, posing significant challenges to fairness. Such bias can lead to real-world harms. For instance, several studies have shown how bias in facial recognition technologies disproportionately misidentifies individuals of certain ethnic backgrounds~\citep{facialSoftwareMisclassification, facialSoftwareMisclassification2}, leading to potential discrimination in law enforcement and hiring practices.

This highlights the urgent need to address AI bias. Ensuring fairness in deep learning models presents complex challenges, primarily due to the black-box nature of these models, which often complicates understanding and interpreting decisions. Moreover, the dynamic and high-dimensional nature of the data involved, combined with nuances in fairness definitions, further complicates the detection and correction of bias. This complexity necessitates the development of more sophisticated, inherently fair algorithms.

Previous mitigation strategies dealing with algorithmic bias--whether through pre-processing, in-processing, or post-processing--have significant limitations. Pre-processing techniques, which attempt to cleanse biased data, are labour-intensive, dependent on expert intervention~\citep{salimi2019interventional}, and only eliminate considered biases. Current in-processing methods frequently lead to unstable models and often rely upon arbitrary definitions of fairness~\citep{fairness-survey}. Post-processing techniques, which adjust model predictions directly, ignore deeper issues without addressing the underlying biases in the data and model. These approaches lack stability, generalisability, and the ability to ensure fairness across multiple metrics~\citep{berk2017convex}.

In this paper, we introduce FairVIC (\textbf{Fair}ness through \textbf{V}ariance, \textbf{I}nvariance, and \textbf{C}ovariance), a novel approach that embeds fairness directly into neural networks by optimising a custom loss function. This function is designed to minimise the correlation between binary decisions and binary protected characteristics while maximising overall prediction performance. FairVIC integrates fairness through the concepts of variance, invariance, and covariance during the training process, making it more principled and intuitive, and universally applicable to diverse datasets. Unlike previous methods that often optimise to a chosen fairness metric, FairVIC offers a robust, generalisable solution that introduces an abstract concept of fairness to significantly reduce bias. Our experimental evaluations demonstrate FairVIC's ability to significantly improve performance in all fairness metrics tested without compromising prediction accuracy. We compare our proposed method against comparable in-processing bias mitigation techniques, such as regularisation and constraint approaches, and highlight the improved, robust performance of our FairVIC model.

Our contributions in this paper are multi-fold:
\begin{itemize}[noitemsep, nolistsep]
    \item A novel, generalisable in-processing bias mitigation technique for neural networks;
    \item A comprehensive experimental evaluation, using a multitude of comparable methods on a variety of metrics across several datasets, including different modalities such as tabular, text, and image;
    \item An extended analysis of our proposed method to examine its robustness, including a full ablation study on the lambda weight terms within our loss function.
\end{itemize}

This paper is structured as follows: Section~\ref{sec:Related Work} discusses current approaches to mitigating bias throughout each processing stage. Section~\ref{sec:Preliminaries} describes any preliminary details for this work, including the fairness metrics used in the evaluation. Section~\ref{sec:Approach} outlines our method, including how each term in our loss function is calculated and an algorithm detailing how these terms are applied. Section~\ref{sec:Experiments} describes the experiments carried out, Section~\ref{sec:Evaluation} outlines the results with discussion, and Section~\ref{sec:Conclusion and Future Work} concludes this work. Extra information, including the dataset metadata and more extensive experiments, is to be found in the Appendix.

\section{Related Work}
\label{sec:Related Work}
There exist three broad categories of mitigation strategies for algorithmic bias: pre-processing, in-processing, and post-processing. Each aims to increase fairness differently by acting upon either the training data, the model itself, or the predictions outputted by the model, respectively.

\textbf{Pre-processing} methods aim to \textit{fix} the data before training, recognising that bias is primarily an issue with the data itself~\citep{fairness-survey}. In practice, this can be done a number of different ways, such as representative sampling, or re-sampling the data to reflect the full population~\citep{Shekhar_Fields_Ghavamzadeh_Javidi_2021, Ustun_Liu_Parkes_2019}, reweighing the data such that different groups influence the model in a representative way~\citep{Calders_Žliobaitė_2013, Kamiran_Calders_2012}, or generating synthetic data to balance out the representation of each group~\citep{jang2021constructing}. Another set of approaches utilises causal methods to delineate relationships between sensitive attributes and the target variables within the data~\citep{chiappa2019causal,kusner2017counterfactual, russell2017worlds}.
Such techniques as these are labour-intensive and do not generalise well, requiring an expert with knowledge of the data to manually process each case of a new dataset~\citep{salimi2019interventional}. They also cannot provide assurances that all bias has been removed--a model may draw upon non-linear/ complex relationships between features that lead to bias, which are hard for the expert/method to spot.

\textbf{In-processing} methods aim to train models to make fairer predictions, even upon biased data. There are a plethora of ways in which this has been done. For example, \citet{celis2019classification} and~\citet{ExponentiatedGradientDescent} utilise a chosen fairness metric and perform constraint optimisation during training. This requires choosing a fairness metric, introducing human bias and limiting generalisation~\cite{fairness-survey}. Therefore, fairness cannot be achieved across multiple definitions in this way~\cite{fairness-survey}. Another approach involves incorporating an adversarial component during model training that penalises the model if protected characteristics can be predicted from its outputs~\citep{zhang2018mitigating, wadsworth2018achieving, xu2019achieving}. These methods are often effective but their main shortcoming is seen in their instability~\cite{han2023ffb}. Finally, the most relevant comparisons from previous work to our proposed method are regularisation-based techniques that incorporate fairness constraints or penalties directly into the model's loss function during training. There are a number of ways that this has been done, such as through data augmentation strategies to promote less sensitive decision boundaries~\citep{FairMixUp} or by incorporating fairness adjustments into the boosting process~\citep{FairGBM}. The performance of these models differs from approach to approach, and those that work by constraining the model by a fairness metric directly suffer from the issue of human bias and misrepresenting the bias within the data/model.

\textbf{Post-processing} techniques involve adjusting model predictions or decision rules after training to ensure fair outcomes. In practice, decision thresholds have been adjusted for different groups to achieve equal outcomes in a particular metric~\citep{EqualOdds}. Alternatively, labels near the decision boundary can be altered to favour less biased outcomes~\citep{kamiran2012decision, kamiran2018exploiting}. Calibration~\citep{Kim_Reingold_Rothblum_2018, Noriega-Campero_Bakker_Garcia-Bulle_Pentland_2019} adjusts the predictions of the model directly so that the proportion of positive instances is equal across each sub-group. These methods oversimplify fairness and ignore model-level bias. For those techniques that require the specification of a single fairness metric, the same issue applies surrounding this choice as before.

To summarise, there currently lies a number of issues which have not yet been solved in parallel within one technique. These are: stability, generalisability, equal improvements to fairness across metrics~\citep{berk2017convex}, and built without requirements for user-induced definitions of fairness. In this paper, we solve all these requirements for an effective, generalisable approach to mitigate bias through FairVIC.


\section{Preliminaries}
\label{sec:Preliminaries}

\subsection{VICReg}
\label{sec:VICReg}
Variance-Invariance-Covariance Regularization (VICReg)~\citep{VICReg} has previously been used in self-supervised learning to tackle feature collapse and redundancy. It maximises variance across features to ensure the model produces diverse outputs for different inputs, minimises invariance between augmented representations of the same input to enhance stability, and reduces covariance among features to capture a broader range of information. \new{VICReg has mostly been confined to self-supervised learning, with little exploration beyond. To this extent, FairVIC reworks the VIC components completely to serve its application in supervised learning for bias mitigation, a principle that has remained so far unexplored.}
This adaptation addresses the challenges of fairness in decision-making systems, expanding the application of VIC principles beyond their original scope and offering a novel, generalisable solution to fairness in supervised learning models.

\subsection{Group Fairness Metrics}
\label{sec:Group Fairness Metrics}
In this section, we introduce notation and state the fairness measures that we use to quantify bias.

\begin{itemize}[leftmargin=0pt, label={}]
    \item \textbf{Equalized Odds Difference} requires that both the True Positive Rate (TPR) and False Positive Rate (FPR) are the same across groups defined by the protected attribute, where $TPR = \frac{TP}{TP + FN}$ and $FPR = \frac{FP}{FP + TN}$~\citep{EqualOdds}. Therefore, we calculate $\max\left(\left| FPR_{u} - FPR_{p} \right|, \left| TPR_{u} - TPR_{p} \right|\right)$, where ${u}$ represents the unprivileged groups and ${p}$ the privileged group and 0 signifies perfect fairness.
    \item \textbf{Average Absolute Odds Difference}  averages the absolute differences in the false positive rates and true positive rates between groups, defined as $\frac{1}{2}(|FPR_{u}-FPR_{p}|+|TPR_{u}-TPR_{p}|)$, where ${u}$ represents the unprivileged groups and ${p}$ the privileged group, with 0 signifying perfect fairness.
    \item \textbf{Statistical Parity Difference}  evaluates the difference in the probability of a positive prediction between groups, aiming for 0 to signify perfect fairness. Formally, $DP=|P(\hat{Y}=1|u)-P(\hat{Y}=1|p)|$, where ${u}$ represents the unprivileged groups, ${p}$--the privileged group, and $\hat{Y}=1$--a positive prediction~\citep{DemographicParity}.
    \item \textbf{Disparate Impact} compares the proportion of positive outcomes for the unprivileged group to that of the privileged group, with a ratio of 1 indicating no disparate impact, and therefore perfect fairness. Denoted as $DI=\frac{P(\hat{Y}=1|u)}{P(\hat{Y}=1|p)}$, where ${u}$ represents the unprivileged groups, ${p}$--the privileged group, and $\hat{Y}=1$--a positive prediction~\citep{DisparateImpact}.
\end{itemize}

\subsection{Individual Fairness}
\label{sec:Individual Fairness}
While FairVIC aims to increase group fairness, the invariance term promotes direct improvements in individual fairness. This can be observed in our evaluations through counterfactual fairness~\cite{kusner2017counterfactual}. Counterfactual fairness ensures that decisions made by an algorithm are fair even when considering hypothetical (counterfactual) scenarios. For each individual, the sensitive attribute is switched to assess the model's ability to perform equally in both the original and counterfactual scenarios.

Formally, if $u$ denotes the unprivileged group, $p$ the privileged group and $\hat{Y}$ is the decision outcome, then the model is considered counterfactually fair if $\hat{Y}_{u} = \hat{Y}_{p}$ for different groups $u$ and $p$ of the sensitive attribute while all non-sensitive features remain the same.


\section{Approach}
\label{sec:Approach}
We propose FairVIC (\textbf{Fair}ness through \textbf{V}ariance, \textbf{I}nvariance, and \textbf{C}ovariance), a novel loss function that enables a model's ability to learn fairness in a robust manner. FairVIC is comprised of three terms: variance, invariance, and covariance. Optimising for these three terms encourages the model to be stable and consistent across protected characteristics, thereby reducing bias during training. This broad, generalised approach to defining bias improves performance across a range of fairness metrics. This makes it an effective strategy for reducing bias across various applications, ensuring more equitable outcomes in diverse settings.

\subsection{FairVIC Training}
\label{sec:fairvic-training}

\new{To understand how FairVIC operates, it is crucial to define \textit{variance}, \textit{invariance}, and \textit{covariance}  in their adapted forms: each is based on a classical statistical concept but used to express a specific fairness objective within the model. We consider binary classification tasks, where the model output \(\hat{y} \in [0, 1]\) represents the predicted probability of the favourable class, obtained via a sigmoid activation.}

\textbf{Variance}: This term promotes diversity in the latent representations by penalising low variance across features in the bottleneck embeddings, \new{denoted \(z\in \mathbb{R}^d \), where \(z\) is the output of the encoder for each input.} It ensures the embeddings capture sufficient information, rather than collapsing to a trivial relationship such as the protected characteristic.

\begin{equation}
    L_{\text{var}} = \frac{1}{d} \sum_{j=1}^{d} \max(0, \gamma - \sigma_j(z))
\label{eq:variance-loss}
\end{equation}
where \(\sigma_j(z)\) represents the standard deviation of the \(j\)-th feature across the batch, we set \(\gamma\) to 1.0 as a margin parameter to encourage unit-scale variability in the latent representation, and \(d\) is the number of features in the embedding.

\textbf{Invariance}: This term ensures the model’s predictions remain consistent when the protected attribute is flipped, promoting individual (counterfactual) fairness.
\begin{equation}
    L_{\text{inv}} = \frac{1}{N} \sum_{i=1}^{N} \left( \hat{y}_i - \hat{y}_i^{*} \right)^2
\label{eq:invariance-loss}
\end{equation}
where $\hat{y}_i$ is the prediction for the original input, $\hat{y}_i^{*} $ is the prediction for the input with the protected attribute $P$ flipped to its complement, and $N$ is the number of samples.

\textbf{Covariance}: This component seeks to \new{penalise any contribution of the protected attribute to the decision process of the classifier, ensuring that predictions are not systematically skewed for the privileged group.} By doing so, it promotes group fairness. The loss function is designed to minimise this covariance, as defined by the following equation:

\begin{equation}
    \new{L_{\text{cov}} = \frac{1}{N} \sqrt{\sum_{i=1}^{N} \left( \left(\hat{y}_i - \mathbb{E}[\hat{y}]\right) \cdot P_i \right)^2}}
\label{eq:covariance-loss}
\end{equation}

\new{where $\hat{y} \in (0,1)$ is the model’s softmax output, $P_i \in \{0,1\}$ is the binary protected attribute with $P_i = 1$ denoting the privileged group, and $N$ is the number of samples. In general, $\mathbb{E}[\hat{y}]$ reflects the empirical average of predictions across the batch, though it may approach 0.5 in a balanced and well-calibrated setting. During training, this loss reduces the deviation of privileged group predictions from the batch mean. In practice, it tends to soften overly confident predictions for the privileged group without necessarily changing the predicted class. This term does not directly affect accuracy, which is controlled by a separate loss, but its interaction with the accuracy objective discourages reliance on group membership in decision-making, thereby promoting group fairness.}

Together, alongside a suitable accuracy loss, FairVIC jointly optimises for the total loss equation seen in Equation~\ref{eq:fairvic-loss}.
\begin{equation}
    \new{L_{\text{total}} = \lambda_{\text{acc}}L_{\text{acc}} + \lambda_{\text{var}}L_{\text{var}} + \lambda_{\text{inv}}L_{\text{inv}} + \lambda_{\text{cov}}L_{\text{cov}}}
\label{eq:fairvic-loss}
\end{equation}
where each $\lambda$ is a non-negative weighting coefficient balancing the contribution of its corresponding term, subject to the normalisation constraint $\lambda_{\text{acc}} + \lambda_{\text{var}} + \lambda_{\text{inv}} + \lambda_{\text{cov}} = 1$.


\section{Experiments}
\label{sec:Experiments}
In our experimental evaluation, we assess the performance of FairVIC\footnote{The code for our FairVIC implementation is provided at: \url{https://github.com/CharmaineBarker/FairVIC}.} against a set of comparable in-processing bias mitigation methods on a series of datasets known for their bias. Here, we describe the datasets used and the methods we compare against.

\subsection{Datasets}
\label{sec:Datasets}
We evaluate FairVIC on seven widely used bias mitigation benchmarks across tabular, text, and image modalities. These datasets contain known demographic disparities, allowing us to assess FairVIC’s generalisability. For each, we designate one attribute as the protected characteristic on which fairness is to be improved.

\begin{itemize}[leftmargin=0pt, label={}, nosep]
    \item \textbf{Tabular datasets.} We use Adult Income~\citep{adult}, COMPAS~\citep{COMPAS}, and German Credit~\citep{german}, all binary classification tasks with known biases. Adult Income predicts whether income exceeds $\$50$K and exhibits gender and racial bias. COMPAS predicts recidivism risk and is notorious for racial bias. German Credit classifies creditworthiness, with biases related to age and gender~\citep{ageThreshold}.
    \item \textbf{Language datasets.} CivilComments-WILDS~\citep{wilds} and BiasBios~\citep{biasbios} are used to assess FairVIC on text data. We sample 50,000 stratified examples per dataset to ensure balance. CivilComments classifies online comments as toxic or non-toxic, using race as the protected attribute. BiasBios consists of professional biographies classified into favourable vs. unfavourable occupations (e.g., surgeon vs. nurse), with gender as the protected feature.
    \item \textbf{Image datasets.} CelebA~\citep{celeba} contains celebrity portraits labelled for attributes such as blond hair; we predict whether a person has blond hair or not, using gender as the protected attribute. UTKFace~\citep{utkface} includes facial images labelled with age and race; we predict whether a subject is above or below 30, with race as the protected attribute.
\end{itemize}

Detailed metadata for each dataset, including our selections for protected groups and classification goals, can be found in Appendix~\ref{sec:Dataset Metadata}.

\subsection{Comparable Techniques}
\label{sec:Comparable Techniques}
To highlight the performance of FairVIC, we evaluate against five comparable in-processing bias mitigation methods. These are:

\begin{itemize}[leftmargin=0pt, label={}, nosep]
    \item \textbf{Adversarial Debiasing.} This method leverages an adversarial network that aims to predict protected characteristics based on the predictions of the main model. The primary model seeks to maximise its own prediction accuracy while minimising the adversary's prediction accuracy~\citep{zhang2018mitigating}. 
    \item \textbf{Exponentiated Gradient Reduction.} This technique reduces fair classification to a sequence of cost-sensitive classification problems, returning a randomised classifier with the lowest empirical error subject to a chosen fairness constraint~\citep{ExponentiatedGradientDescent}.
    \item \textbf{Meta Fair Classifier.} This classifier allows a fairness metric as an input and optimises the model with respect to regular performance and the chosen fairness metric~\citep{celis2019classification}.
    \item \textbf{Fair MixUp.} This technique generates synthetic samples by linearly interpolating between pairs of training data points by protected attribute to smooth decision boundaries. The loss function is then further constrained by a fairness metric~\citep{FairMixUp}.
    \item \textbf{FairGBM.} This method uses a gradient-boosting decision tree model that integrates fairness constraints directly into the boosting process by adjusting the loss function to account for fairness metrics~\citep{FairGBM}.
\end{itemize}

A baseline neural network trained with binary cross-entropy was also implemented to reflect dataset biases. Model architecture and hyperparameters for both baseline and FairVIC are detailed in Appendix~\ref{sec:Neural Network Configuration}.


\section{Evaluation}
\label{sec:Evaluation}

\subsection{Core Results Analysis}
\label{sec:Core Results}
To assess the prediction and fairness performance of FairVIC\footnote{See Table~\ref{tab:Final Lambdas}, Appendix~\ref{sec:Hyperparameter Recommendations} for our FairVIC lambdas, and discussion on these selections.} and state-of-the-art approaches, we test all methods across each tabular dataset to enable a fair comparison. Table~\ref{tab:CoreResults} shows these results. We have also provided Figure~\ref{fig:full-vis}, which visualises the absolute difference from the ideal value of each metric, highlighting how far each method deviates from perfect accuracy and fairness on each tabular dataset.

\begin{table}[!h]
\caption{Accuracy and fairness results of FairVIC versus the baseline and five in-processing methods on three tabular datasets. \textbf{Bold} indicates the best overall trade-off between fairness and accuracy, reflecting our aim of holistic improvement (see Figure~\ref{fig:full-vis}). \underline{Underlined} values show the best score for each individual metric.}
\label{tab:CoreResults}
\resizebox{\textwidth}{!}{%
\begin{tabular}{clllcccc}
\toprule
\multicolumn{2}{c}{} &
\multicolumn{2}{c}{Performance Metrics} &
\multicolumn{4}{c}{Fairness Metrics} \\
\cmidrule(lr){3-4} \cmidrule(lr){5-8}
\multicolumn{1}{c}{Dataset} &
\multicolumn{1}{c}{Model} &
\multicolumn{1}{c}{Accuracy} &
\multicolumn{1}{c}{F1 Score} &
\multicolumn{1}{c}{Equalized Odds} &
\multicolumn{1}{c}{Absolute Odds} &
\multicolumn{1}{c}{Statistical Parity} &
\multicolumn{1}{c}{Disparate Impact} \\
\midrule
&
  Baseline (Biased) &
   0.8444 $\pm$ 0.0065&
   0.6685 $\pm$ 0.0118&
   0.1330 $\pm$ 0.0317&
   0.1172 $\pm$ 0.0289&
   -0.2173 $\pm$ 0.0291&
   0.2853 $\pm$ 0.0329\\  
&
  Adversarial Debiasing &
  0.8065 $\pm$ 0.0048 &
  0.4773 $\pm$ 0.0708 &
  0.2127 $\pm$ 0.0828 &
  0.1172 $\pm$ 0.0443 &
  -0.0405 $\pm$ 0.0679 &
  0.7874 $\pm$ 0.2185 \\  
&
  Exponentiated Gradient Reduction &
  0.8027 $\pm$ 0.0026 &
  0.4056 $\pm$ 0.0052 &
  \underline{0.0238 $\pm$ 0.0115}&
  \underline{0.0167 $\pm$ 0.0061}&
  -0.0601 $\pm$ 0.0026 &
  0.4602 $\pm$ 0.0237 \\  
&
  Meta Fair Classifier &
  0.5171 $\pm$ 0.0602 &
  0.4744 $\pm$ 0.0219 &
  0.4826 $\pm$ 0.0894 &
  0.2935 $\pm$ 0.0497 &
  -0.2098 $\pm$ 0.0542 &
  0.7140 $\pm$ 0.0812 \\  
&
  Fair MixUp  &
  0.7785 $\pm$ 0.0069 &
  0.3815 $\pm$ 0.0521 &
  0.1137 $\pm$ 0.0928 &
  0.0786 $\pm$ 0.0649 &
  -0.0859 $\pm$ 0.0591 &
  0.4830 $\pm$ 0.2174 \\  
&
  FairGBM &
  \underline{0.8731 $\pm$ 0.0026}&
  \underline{0.7122 $\pm$ 0.0079}&
  0.0658 $\pm$ 0.0131 &
  0.0583 $\pm$ 0.0092 &
  -0.1707 $\pm$ 0.0044 &
  0.3363 $\pm$ 0.0151 \\  
\multirow{-7}{*}{\makecell{Adult\\Income}} &
  FairVIC &
  \textbf{0.8284 $\pm$ 0.0088}&
  \textbf{0.5314 $\pm$ 0.0509}&
  \textbf{0.2993 $\pm$ 0.0683}&
  \textbf{0.1637 $\pm$ 0.0371}&
  \textbf{\underline{-0.0088 $\pm$ 0.0249}}&
  \textbf{\underline{0.9803 $\pm$ 0.2220}}\\
\midrule
 &
  Baseline (Biased) &
   \underline{0.6622 $\pm$ 0.0150}&
   0.6118 $\pm$ 0.0252&
   0.3281 $\pm$ 0.0574&
   0.2635 $\pm$ 0.0452&
   -0.2941 $\pm$ 0.0459&
   0.6223 $\pm$ 0.0504\\  
 &
  Adversarial Debiasing &
  0.6581 $\pm$ 0.0185 &
  0.6253 $\pm$ 0.0124 &
  0.1707 $\pm$ 0.0694 &
  0.1363 $\pm$ 0.0504 &
  -0.0902 $\pm$ 0.1367 &
  0.8982 $\pm$ 0.2614 \\  
 &
  Exponentiated Gradient Reduction &
  0.5574 $\pm$ 0.0169 &
  0.2981 $\pm$ 0.0407 &
  \underline{0.0630 $\pm$ 0.0333}&
  \underline{0.0432 $\pm$ 0.0231}&
  \underline{-0.0393 $\pm$ 0.0257}&
  \underline{0.9545 $\pm$ 0.0293}\\  
 &
  Meta Fair Classifier &
  0.3471 $\pm$ 0.0147 &
  0.4312 $\pm$ 0.0380 &
  0.2951 $\pm$ 0.1038 &
  0.2257 $\pm$ 0.1095 &
  0.2526 $\pm$ 0.1070 &
  2.5876 $\pm$ 0.6627 \\  
 &
  Fair MixUp &
  0.6122 $\pm$ 0.0191 &
  0.5356 $\pm$ 0.0437 &
  0.1180 $\pm$ 0.0774 &
  0.0871 $\pm$ 0.0597 &
  -0.0496 $\pm$ 0.0998 &
  0.9427 $\pm$ 0.1470 \\  
 &
  FairGBM &
  0.6440 $\pm$ 0.0151&
  \underline{0.6254 $\pm$ 0.0153}&
  0.2015 $\pm$ 0.1128 &
  0.1466 $\pm$ 0.0961 &
  0.0881 $\pm$ 0.1225 &
  1.2828 $\pm$ 0.4058 \\  
\multirow{-7}{*}{COMPAS} &
  FairVIC &
  \textbf{0.6501 $\pm$ 0.0173} &
  \textbf{0.5934 $\pm$ 0.0357} &
  \textbf{0.0976 $\pm$ 0.0375} &
  \textbf{0.0719 $\pm$ 0.0305} &
  \textbf{-0.0602 $\pm$ 0.0678} &
  \textbf{0.9139 $\pm$ 0.1135} \\
\midrule
&
  Baseline (Biased) &
   0.7255 $\pm$ 0.0284&
   0.8077 $\pm$ 0.0275&
   0.2234 $\pm$ 0.0974&
   0.1641 $\pm$ 0.0936&
   -0.2218 $\pm$ 0.0901&
   0.7140 $\pm$ 0.1203\\  
&
  Adversarial Debiasing &
  0.5815 $\pm$ 0.1513 &
  0.6302 $\pm$ 0.2581 &
  0.1020 $\pm$ 0.0418 &
  0.0737 $\pm$ 0.0404 &
  -0.0657 $\pm$ 0.0335 &
  0.8084 $\pm$ 0.2130 \\  
&
  Exponentiated Gradient Reduction &
  0.7465 $\pm$ 0.0300 &
  \underline{0.8321 $\pm$ 0.0208}&
  0.1232 $\pm$ 0.0631 &
  0.0796 $\pm$ 0.0348 &
  -0.1084 $\pm$ 0.0746 &
  0.8692 $\pm$ 0.0896 \\  
&
  Meta Fair Classifier &
  \underline{0.7575 $\pm$ 0.0260}&
  0.8291 $\pm$ 0.0229 &
  0.2215 $\pm$ 0.1112 &
  0.1444 $\pm$ 0.0810 &
  -0.1052 $\pm$ 0.1315 &
  0.8601 $\pm$ 0.1755 \\  
&
  Fair MixUp &
  0.6925 $\pm$ 0.0225 &
  0.7837 $\pm$ 0.0208 &
  \underline{0.0661 $\pm$ 0.0389}&
  \underline{0.0465 $\pm$ 0.0252}&
  -0.0461 $\pm$ 0.0446 &
  0.9347 $\pm$ 0.0629 \\  
&
  FairGBM &
  0.7460 $\pm$ 0.0348 &
  0.8255 $\pm$ 0.0283 &
  0.1922 $\pm$ 0.0906 &
  0.1345 $\pm$ 0.0756 &
  -0.1539 $\pm$ 0.0773 &
  0.8081 $\pm$ 0.0915 \\  
\multirow{-6}{*}{\makecell{German\\Credit}} &
  FairVIC &
  \textbf{0.7250 $\pm$ 0.0239} &
  \textbf{0.8108 $\pm$ 0.0237} &
  \textbf{0.1443 $\pm$ 0.0796} &
  \textbf{0.1017 $\pm$ 0.0464} &
  \underline{\textbf{0.0016 $\pm$ 0.0604}}&
  \underline{\textbf{1.0037 $\pm$ 0.0764}}\\
\bottomrule
\end{tabular}
}
\end{table}

Across all three datasets, the baseline performs poorly in fairness but obtains higher performance scores, which is expected. For example, in the Adult Income dataset, the baseline model shows a relatively high accuracy ($0.8444$), while exhibiting poor fairness with regard to Disparate Impact ($0.2853$). The baseline highlights the need for a bias mitigation approach that works across all metrics simultaneously, as the lower bias in terms of Equalised Odds ($0.1330$) and Absolute Odds ($0.1172$) alone could misleadingly suggest that the model is fair, when in reality, the bias may only become evident when captured through a different perspective. Single-metric approaches like EGR often fail to address significant bias.

\begin{figure}[tbh!]
\begin{center}
    \includegraphics[width=0.9\textwidth]{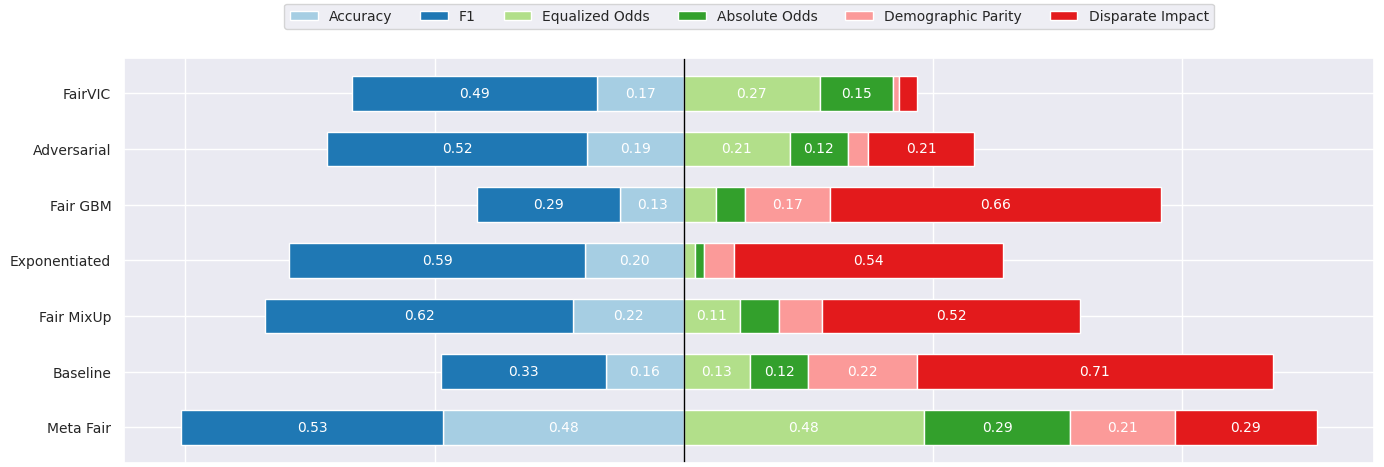}
    \caption{A qualitative analysis of the absolute differences from the ideal value (e.g., perfect accuracy and fairness) in performance (left) and fairness (right) metrics of comparable techniques on the Adult Income dataset.}
    \label{fig:adult-vis}
\end{center}
\end{figure}

Overall, FairVIC outperforms all other comparable methods by demonstrating consistent improvements in both fairness and accuracy retention. As seen in Figure~\ref{fig:adult-vis}, our FairVIC model achieves the lowest cumulative absolute error from perfect accuracy and fairness in the Adult Income dataset, effectively balancing the fairness-accuracy trade-off. The trend is also consistent across the COMPAS and German Credit datasets seen in Figure~\ref{fig:full-vis}, Appendix~\ref{sec:Full Training Results}. This further highlights FairVIC’s ability to generalise across datasets.

Other comparable methods are generally not as effective as FairVIC, each exhibiting different shortcomings. For instance, MetaFair often struggles to improve even upon the baseline in cumulative absolute difference from the ideal value, and many techniques struggle to balance the improvements across all fairness metrics, often prioritising Equalised and Absolute Odds over Disparate Impact, particularly in the Adult Income dataset. Similarly, FairMixUp, though initially promising and achieving second place after FairVIC in the COMPAS and German Credit datasets, fails to maintain its performance on the Adult Income dataset, where its results only just beat the baseline. In many cases, such as FairMixUp on the COMPAS and German Credit datasets, comparable techniques improve fairness but at the cost of accuracy, failing to achieve a balanced tradeoff. Exponentiated Gradient Reduction (EGR) sees similar struggles. While performing the best in terms of each individual fairness metric in the COMPAS dataset, it does this at the expense of accuracy, where it sees a drop of $15.83\%$ in accuracy. Further to this, EGR then performs poorly on the Adult Income dataset, only achieving a disparate impact of 0.4602, suggesting inability to generalise across datasets.

Following this paper’s objective to create an approach that performs well across multiple fairness metrics without significantly compromising accuracy, we find that FairVIC demonstrates a consistent ability to balance fairness and accuracy across diverse datasets. Its adaptability, strong performance on all fairness metrics, and robustness to dataset shifts position it as the most effective method overall. This is further supported by its consistently low cumulative absolute error across both fairness and performance measures, highlighting its advantage over existing in-processing techniques.

\subsection{Individual Fairness Analysis}
\label{sec:Individual Fairness Analysis}
To emphasise further FairVIC's ability to perform well across all fairness metrics, we also evaluate upon individual fairness by outputting the results of the counterfactual model, as described in Section~\ref{sec:Individual Fairness}. The full results, alongside the absolute difference in averages for each metric across the regular and counterfactual models, are seen in Table~\ref{tab:ind-results}, Appendix~\ref{sec:Individual Fairness Results}.

The FairVIC model shows considerable promise in enhancing individual fairness across different datasets when compared with the baseline models. The counterfactual results from the FairVIC model with invariance term weighted heavily (FairVIC Invariance) exhibits lower absolute differences in metrics across all datasets. For example, in the German Credit dataset, the mean absolute difference across all six metrics between the regular and the counterfactual baseline model is $0.0277$, while for FairVIC Invariance's regular and counterfactual models it is lower at $0.0108$. This suggests a more stable and fair performance under counterfactual conditions. This capability highlights FairVIC's strength in not only addressing group fairness but also ensuring that individual decisions remain consistent and fair when hypothetical scenarios are considered. In the FairVIC model with recommended lambdas, we prioritise group fairness so invariance is weighted less. Even with this lower invariance weighting, FairVIC still achieved improved individual fairness.

\begin{table}[!thb]
\caption{FairVIC and baseline comparison results of both performance and fairness for each of the four text and image datasets, including FairVIC's counterfactual (CF) model results and the absolute differences (ADs) between each model.}
\label{tab:LanguageResults}
\centering
\resizebox{0.9\textwidth}{!}{%
\begin{tabular}{clllcccc}
\toprule
\multicolumn{2}{c}{} &
\multicolumn{2}{c}{Performance Metrics} &
\multicolumn{4}{c}{Fairness Metrics} \\
\cmidrule(lr){3-4} \cmidrule(lr){5-8}
\multicolumn{1}{c}{Dataset} &
\multicolumn{1}{c}{Model} &
\multicolumn{1}{c}{Accuracy} &
\multicolumn{1}{c}{F1 Score} &
\multicolumn{1}{c}{Equalized Odds} &
\multicolumn{1}{c}{Absolute Odds} &
\multicolumn{1}{c}{Statistical Parity} &
\multicolumn{1}{c}{Disparate Impact} \\
\midrule
& Baseline &
   0.7624 $\pm$ 0.0055&
   0.7566 $\pm$ 0.0091&
   0.3095 $\pm$ 0.0297&
   0.1832 $\pm$ 0.0236&
   0.2639 $\pm$ 0.0212&
   1.9390 $\pm$ 0.1135\\
 & Baseline CF &
   0.7608 $\pm$ 0.0031&
   0.7608 $\pm$ 0.0069&
   0.3104 $\pm$ 0.0296&
   0.1848 $\pm$ 0.0222&
   -0.2648 $\pm$ 0.0199&
   0.4940 $\pm$ 0.0400\\ 
 & Baseline AD &
   0.0016 $\pm$ 0.0041&
   0.0041 $\pm$ 0.0081&
   0.0009 $\pm$ 0.0110&
   0.0016 $\pm$ 0.0091&
   0.5287 $\pm$ 0.0348&
   1.4449 $\pm$ 0.1360\\ 
 & FairVIC &
   0.7243 $\pm$ 0.0755&
   0.6613 $\pm$ 0.1954&
   0.1457 $\pm$ 0.0661&
   0.1030 $\pm$ 0.0429&
   0.0562 $\pm$ 0.0517&
   1.1344 $\pm$ 0.1452\\ 
 & FairVIC CF &
   0.6323 $\pm$ 0.1057&
   0.5722 $\pm$ 0.2128&
   0.1316 $\pm$ 0.0846&
   0.0953 $\pm$ 0.0819&
   0.0233 $\pm$ 0.1006&
   1.0687 $\pm$ 0.2324\\ 
 \multirow{-6}{*}{\makecell{CivilComments\\-WILDS}} & FairVIC AD &
   0.0921 $\pm$ 0.0907&
   0.0892 $\pm$ 0.2381&
   0.0141 $\pm$ 0.0711&
   0.0077 $\pm$ 0.0751&
   0.0329 $\pm$ 0.0648&
   0.0657 $\pm$ 0.1844\\
\midrule
\multirow{6}{*}{\makecell{BiasBios}} & Baseline &
0.8818 $\pm$ 0.0034&
   0.8811 $\pm$ 0.0044&
   0.0558 $\pm$ 0.0103&
   0.0456 $\pm$ 0.0083&
   -0.2489 $\pm$ 0.0098&
   0.6038 $\pm$ 0.0159\\ 
 & Baseline CF &
   0.8794 $\pm$ 0.0041&
   0.8797 $\pm$ 0.0032&
   0.0563 $\pm$ 0.0153&
   0.0461 $\pm$ 0.0120&
   0.2481 $\pm$ 0.0119&
   1.6401 $\pm$ 0.0292\\ 
 & Baseline AD &
   0.0041 $\pm$ 0.0029&
   0.0037 $\pm$ 0.0024&
   0.0123 $\pm$ 0.0093&
   0.0066 $\pm$ 0.0042&
   0.4970 $\pm$ 0.0206&
   1.0363 $\pm$ 0.0421\\ 
 & FairVIC &
    0.8668 $\pm$ 0.0070&
    0.8633 $\pm$ 0.0087&
    0.0970 $\pm$ 0.0167&
    0.0830 $\pm$ 0.0084&
    -0.1220 $\pm$ 0.0110&
    0.7744 $\pm$ 0.0207\\ 
 & FairVIC CF &
    0.8601 $\pm$ 0.0069&
    0.8582 $\pm$ 0.0059&
    0.1464 $\pm$ 0.0222&
    0.1181 $\pm$ 0.0112&
    0.0862 $\pm$ 0.0117&
    1.1945 $\pm$ 0.0304\\ 
 & FairVIC AD &
    0.0066 $\pm$ 0.0063&
    0.0051 $\pm$ 0.0070&
    0.0493 $\pm$ 0.0240&
    0.0350 $\pm$ 0.0146&
    0.2082 $\pm$ 0.0132&
    0.4201 $\pm$ 0.0295\\
\midrule
 & Baseline &
    0.8765 $\pm$ 0.0048&
    0.8776 $\pm$ 0.0057&
    0.2019 $\pm$ 0.0516&
    0.1268 $\pm$ 0.0249&
    -0.4672 $\pm$ 0.0124&
    0.4353 $\pm$ 0.0208\\ 
 & Baseline CF &
    0.8747 $\pm$ 0.0063&
    0.8724 $\pm$ 0.0106&
    0.2286 $\pm$ 0.0495&
    0.1396 $\pm$ 0.0232&
    0.4676 $\pm$ 0.0096&
    2.2250 $\pm$ 0.1288\\ 
 & Baseline AD &
    0.0018 $\pm$ 0.0055&
    0.0052 $\pm$ 0.0073&
    0.0267 $\pm$ 0.0510&
    0.0127 $\pm$ 0.0246&
    0.9348 $\pm$ 0.0176&
    1.7897 $\pm$ 0.1319\\ 
 & FairVIC &
    0.7571 $\pm$ 0.0286&
    0.7811 $\pm$ 0.0249&
    0.2788 $\pm$ 0.0869&
    0.1543 $\pm$ 0.0349&
    -0.1018 $\pm$ 0.0815&
    0.8259 $\pm$ 0.1596\\ 
 & FairVIC CF &
    0.7442 $\pm$ 0.0140&
    0.7804 $\pm$ 0.0180&
    0.3189 $\pm$ 0.0372&
    0.1686 $\pm$ 0.0148&
    0.0612 $\pm$ 0.0318&
    1.1956 $\pm$ 0.0909 \\ 
\multirow{-6}{*}{\makecell{CelebA}} & FairVIC AD &
   0.0129 $\pm$ 0.0142 &
   0.0006 $\pm$ 0.0239 &
   0.0401 $\pm$ 0.0584 &
   0.0142 $\pm$ 0.0214 &
   0.1630 $\pm$ 0.0958 &
   0.3697 $\pm$ 0.1949 \\
\midrule
\multirow{6}{*}{\makecell{UTKFace}} & Baseline &
   0.6827 $\pm$ 0.0120&
   0.6839 $\pm$ 0.0225&
   0.0887 $\pm$ 0.0322&
   0.0631 $\pm$ 0.0260&
   0.1261 $\pm$ 0.0265&
   1.3136 $\pm$ 0.1024\\ 
 & Baseline CF &
   0.6723 $\pm$ 0.0230&
   0.6655 $\pm$ 0.0767&
   0.0988 $\pm$ 0.0218&
   0.0741 $\pm$ 0.0224&
   -0.1328 $\pm$ 0.0233&
   0.7416 $\pm$ 0.0868\\ 
 & Baseline AD &
   0.0104 $\pm$ 0.0198&
    0.0184 $\pm$ 0.0521&
   0.0100 $\pm$ 0.0106&
   0.0110 $\pm$ 0.0101&
   0.2589 $\pm$ 0.0450&
   0.5720 $\pm$ 0.1806\\ 
 & FairVIC &
   0.6308 $\pm$ 0.0288&
   0.5901 $\pm$ 0.1011&
   0.0647 $\pm$ 0.0335&
   0.0509 $\pm$ 0.0313&
   -0.0035 $\pm$ 0.0332&
   1.0076 $\pm$ 0.0770\\ 
 & FairVIC CF &
   0.6231 $\pm$ 0.0247&
   0.5713 $\pm$ 0.1176&
   0.0796 $\pm$ 0.0367&
   0.0645 $\pm$ 0.0325&
   0.0146 $\pm$ 0.0385&
   1.0097 $\pm$ 0.1332\\ 
 & FairVIC AD &
   0.0077 $\pm$ 0.0250&
   0.0188 $\pm$ 0.1106&
   0.0149 $\pm$ 0.0317&
   0.0136 $\pm$ 0.0251&
   0.0181 $\pm$ 0.0334&
   0.0021 $\pm$ 0.0894\\
   \bottomrule
\end{tabular}
}
\end{table}

\subsection{Language and Image Dataset Results}
\label{sec:Language and Image Dataset Results}
To demonstrate FairVIC's versatility, we apply it to CivilComments-WILDS and BiasBios. Results are shown in Table~\ref{tab:LanguageResults}; our selected lambda configurations are given in Table~\ref{tab:Final Lambdas}. Across all fairness metrics, FairVIC improves upon the baseline, consistent with trends seen in the tabular datasets. Notably, Disparate Impact improves from $1.9390$ to $1.1344$ (CivilComments-WILDS), $0.6038$ to $0.7817$ (BiasBios), $0.4353$ to $0.8259$ (CelebA), and $1.3136$ to $1.0076$ (UTKFace). For individual fairness, the mean absolute difference between regular and counterfactual models is reduced substantially: from $0.3303$ to $0.0503$ on CivilComments-WILDS, $0.2563$ to $0.1207$ on BiasBios, $0.4618$ to $0.1001$ on CelebA, and $0.1468$ to $0.0125$ on UTKFace. These results confirm FairVIC’s effectiveness across modalities and architectures, improving both group and individual fairness with minimal accuracy trade-off.

\subsection{Lambda Ablation Study Analysis}
\label{sec:Lambda Ablation Study Analysis}
Finally, to better understand the internal dynamics of FairVIC, we conduct an ablation study on the $\lambda$ weightings of FairVIC’s loss terms to analyse their individual contributions. The FairVIC loss terms are combined with binary cross entropy for training the neural network to enable optimisation of both accuracy and fairness, minimising the trade-off. The effect of FairVIC on the overall loss function can be increased and decreased by changing the weight $\lambda$ for each FairVIC term. To evaluate this effect, we train a number of neural networks with the architecture described in Appendix~\ref{sec:Neural Network Configuration}, with a different $\lambda_{\text{acc}}$ weighting each time. In this experiment, we evaluate the effect of weighting the FairVIC loss terms equally, so that $\lambda_{\text{var}}=\lambda_{\text{inv}}=\lambda_{\text{cov}}=\frac{(1-\lambda_{\text{acc}})}{3}$, where $0<\lambda_{\text{acc}}<1$. The performance and fairness measures for each model are listed in Table~\ref{tab:equal_vic_results}, Appendix~\ref{sec:Lambda Ablation Study Results}, and visualisations for the absolute difference in performance and fairness from ideal values for each run are visualised in Figure~\ref{fig:trade-off-graph}, Appendix~\ref{sec:Lambda Ablation Study Results}.

In Figure~\ref{fig:trade-off-graph} (Appendix~\ref{sec:Lambda Ablation Study Results}), the trade-off between accuracy and fairness is evident. As $\lambda_{\text{acc}}$ increases, predictive performance improves, but the fairness metrics deviate further from the ideal value. In contrast, when $\lambda_{\text{acc}}$ is lower, fairness improves, but this time with only a negligible drop in accuracy. This suggests that lower $\lambda_{\text{acc}}$ values provide a better overall performance balance. This trend is much more prevalent for the larger Adult dataset, where more complex relationships could lead to a larger accuracy-fairness trade-off. In the COMPAS and German Credit datasets, this trade-off, while still following the same pattern, is much smaller.

To evaluate the effect of each individual VIC term within the loss function, we can suppress the lambda terms from two out of three of variance, invariance, and covariance to leave only one remaining. We keep $\lambda_{\text{acc}}=0.1$ since the previous lambda experiment showed this to be most effective and revealing in terms of the effect on fairness, while the chosen FairVIC loss term is assigned a weighting of $0.9$. Similarly, we can also suppress a single term at a time, assigning two out of the three VIC terms a weighting of $0.45$. The performance and fairness results for each experiment with different weightings are listed in Table~\ref{tab:individual_vic_results}.

It can be concluded that each term has a different effect. The variance term is shown to have the lowest standard deviation across all metrics and all tabular data in Table~\ref{tab:individual_vic_results}, offering stability to FairVIC. The covariance term makes the greatest contribution to group fairness, as seen in Table~\ref{tab:individual_vic_results}. The invariance term aims to give similar outputs to similar inputs, regardless of the protected attribute; therefore, it should have more of an effect towards individual fairness. Table~\ref{tab:ind-results} corroborates this hypothesis, as the FairVIC Invariance model (FairVIC with the invariance loss term weighted to $0.9$, and accuracy loss of $0.1$) consistently has a lower absolute difference than the baseline between the regular and counterfactual models across all metrics and tabular datasets, signalling greater individual fairness. Therefore, we conclude that the combination of all three terms would aim to improve both group and individual fairness, and increase stability.

Based on these findings, we can offer our recommendations for default lambda values to achieve effective results. We suggest beginning with a configuration of $\lambda_{\text{acc}} = 0.1$, $\lambda_{\text{var}} = 0.1$, $\lambda_{\text{inv}} = 0.1$, and $\lambda_{\text{cov}} = 0.7$, which consistently performs well across datasets. This configuration provides a strong default, with further tuning advised based on the relative importance of group fairness, individual fairness, or accuracy in a given application. A detailed discussion of dataset-specific configurations and practical guidance for adjusting these weights is provided in Appendix~\ref{sec:Hyperparameter Recommendations}.

\section{Conclusion and Future Work}
\label{sec:Conclusion and Future Work}
In this paper, we introduced FairVIC, an in-processing bias mitigation technique that introduces three new terms into the loss function of a neural network- variance, invariance, and covariance. Across our experimental evaluation, FairVIC significantly improves scores for all fairness metrics, with minimal drop in accuracy, compared to previous comparable methods which typically aim to improve only upon a single metric. This balance showcases FairVIC’s strength in providing a robust and effective solution applicable across various tasks and datasets. Future work will extend FairVIC to multiple protected attributes and targets.

\section*{Declaration on Generative AI}
The author(s) have not employed any Generative AI tools.

\bibliography{sample-ceur}

\newpage
\appendix
\onecolumn
\section{Experiment Details}
\label{sec:Experiment Details}

\subsection{FairVIC Training Algorithm}
\label{sec:FairVIC Training Algorithm}
During training, the model iterates over epochs $E$, with the data shuffled into batches. For each batch, the model produces predictions $\hat{Y}$, which are compared with the true labels $Y$ using a suitable accuracy loss function (e.g., binary cross-entropy, hinge loss, or Huber loss). The resulting loss is then minimised by an optimiser.

The pseudocode below summarises the full training procedure for FairVIC. It integrates the accuracy loss alongside the proposed fairness-promoting objectives- variance, invariance, and covariance- as defined in Section~\ref{sec:fairvic-training}. The total loss is computed as a weighted combination of these terms. Subsequently, gradients are computed, and the optimiser adjusts the model parameters with respect to this combined loss.

The multipliers $\lambda$ enable users to balance the trade-off between fairness and predictive performance, which is typical in bias mitigation techniques. Assigning a higher weight to $\lambda_{\text{acc}}$ directs the model to prioritise accuracy, while increasing the weights of $(\lambda_{\text{var}}, \lambda_{\text{inv}}, \lambda_{\text{cov}})$ shifts the focus towards enhancing fairness in the model’s predictions. In our implementation, the lambda coefficients \((\lambda_{\text{acc}}, \lambda_{\text{var}}, \lambda_{\text{inv}}, \lambda_{\text{cov}})\) are constrained such that their sum equals one. In other words, $\lambda_{\text{acc}} = 1 - \lambda_{\text{var}} - \lambda_{\text{inv}} - \lambda_{\text{cov}}$. This normalisation ensures the optimisation will not produce multiple solutions in the form $\{k \cdot \lambda_{\text{acc}}, k \cdot \lambda_{\text{var}}, k \cdot \lambda_{\text{inv}}, k \cdot \lambda_{\text{cov}}\}, \; k \in \mathbb{R}$.

\begin{center}
\begin{varwidth}{0.8\textwidth}
\begin{algorithm}[H]
\caption{FairVIC Loss Function}
\label{alg:fairvic-loss}
\begin{algorithmic}[1]
\State \textbf{Input:} Model $M$, Epochs $E$, Batch size $B$, Data $D$, Protected attribute $P$, Weights ($\lambda_{\text{acc}}, \lambda_{\text{var}}, \lambda_{\text{inv}}, \lambda_{\text{cov}}$)
\State \textbf{Output:} Trained Model $M$
\State Initialise $M$
\For{$e \in \{1, \dots, E\}$}
\State Shuffle data $D$
    \For{each batch $\{(X, Y)\} \in D$ with size $B$}
        \State $\hat{Y} \gets M(X)$
        \State $Z \gets \text{BottleneckLayer}(X)$

        \State Calculate FairVIC Loss:
        \State \ \ \ $L_{\text{acc}} \gets \text{AccuracyLoss}(Y, \hat{Y})$
        \State \ \ \ $L_{\text{var}} \gets \text{VarianceLoss}(Z)$
        \State \ \ \ $L_{\text{inv}} \gets \text{InvarianceLoss}(\hat{Y}, M(\text{Flip}(X, P)))$
        \State \ \ \ $L_{\text{cov}} \gets \text{CovarianceLoss}(\hat{Y}, P)$
        \State \ \ \ $L_{\text{total}} \gets \lambda_{\text{acc}} L_{\text{acc}} + \lambda_{\text{var}} L_{\text{var}} + \lambda_{\text{inv}} L_{\text{inv}} + \lambda_{\text{cov}} L_{\text{cov}}$

        \State Compute gradients $\nabla L_{\text{total}} \gets \frac{\partial L_{\text{total}}}{\partial M}$
        \State Update model parameters $M \gets M - \alpha \nabla L_{\text{total}}$
    \EndFor
\EndFor
\end{algorithmic}
\end{algorithm}
\end{varwidth}
\end{center}

\subsection{Dataset Metadata}
\label{sec:Dataset Metadata}
Detailed metadata for each dataset, including our selection of privileged group, can be found in Table~\ref{tab:Datasets}. Note that for the language datasets, the number of features is obtained by combining the protected characteristic and the target label with the 50 tokenised text features and for image datasets, this is the pixels in an image plus the protected attribute and target label. For BiasBios, we take architect, attorney, dentist, physician, professor, software engineer, surgeon as the favourable professions, and interior designer, journalist, model, nurse, poet, teacher, and yoga teacher as the unfavourable professions for our binary classification task.

\begin{table}[htb]
\caption{Metadata on all seven experimental datasets.}
\label{tab:Datasets}
\begin{center}
\resizebox{0.9\columnwidth}{!}{%
\begin{tabular}{llllllll}
\toprule
\multicolumn{1}{c}{Dataset} &
\multicolumn{1}{c}{Adult Income} &
\multicolumn{1}{c}{COMPAS} & 
\multicolumn{1}{c}{German Credit} & 
\multicolumn{1}{c}{CivilComments-WILDS} & 
\multicolumn{1}{c}{BiasBios} & 
\multicolumn{1}{c}{CelebA} & 
\multicolumn{1}{c}{UTKFace} \\ \midrule
Data Modality            & Tabular               & Tabular              & Tabular              & Text            & Text              &Image &Image \\
No. of Features          & 11                    & 8                    & 20                   & 52              & 52                & 128x128 +2 & 128x128 +2 \\
No. of Rows              & 48,842                & 5,278                & 1,000                & 50,000          & 50,000            & 50,000 & 23,705 \\
\midrule
Target Variable          & income                & two\_year\_recid     & credit               & toxicity        & profession        & Blond$\_$Hair & age \\
Favourable Label         & \textgreater{}50K (1) & False (0)            & Good (1)             & Non-Toxic (0)   & Favourable (1)    & Not Blond (-1) & \textless{}30 (1) \\
Unfavourable Label       & \textless{}=50K (0)   & True (1)             & Bad (0)              & Toxic (1)       & Unfavourable (0)  & Blond (1) & \textgreater{}=30 (0) \\
\midrule
Protected Characteristic & sex                   & race                 & age                  & race            & gender            & Male & race \\
Privileged Group         & male (1)              & Caucasian (1)        & \textgreater{}25 (1) & white (1)       & Male (0)          & Male (1) & White (1) \\
Unprivileged Group       & female (0)            & African-American (0) & \textless{}=25 (0)   & non-white (0)   & Female (1)        & Female (0) & Non-White (0) \\
\bottomrule
\end{tabular}
}
\end{center}
\end{table}

\subsection{Neural Network Configuration}
\label{sec:Neural Network Configuration}
The configurations for the neural networks utilised for the tabular, language, and image data can be seen in Table~\ref{tab:Architecture}. To obtain results, each model was run 10 times over random seeds, with a randomised train/test split each time. The averages and standard deviations were then outputted from across all 10 of the runs.

\begin{table}[htb]
\caption{Experimental model setup and parameters.}
\label{tab:Architecture}
\begin{center}
\resizebox{0.9\columnwidth}{!}{%
\begin{tabular}{llllll}
\toprule
\multicolumn{1}{c}{Parameter} & \multicolumn{1}{c}{Tabular Datasets} & \multicolumn{1}{c}{Language Datasets} & \multicolumn{1}{c}{Image Datasets} \\
\midrule
Network Architecture & Dense(128, 64, 32, 2, 32, 64, 128) & BiLSTM(64,32) + Dense(64, 2, 64) & Conv2D(32, 64)+ Dense(128, 64, 32, 2, 32, 64, 128) \\
No. of Epochs & 200 & 50 & 50 \\
Batch Size & 256 & 256 & 256 \\
Optimiser & Adam & AdamW & AdamW \\
Learning Rate & 5e-2 & 5e-5 & 5e-5 \\
Dropout Rate & 0.25 & 0.50 & 0.25 \\
Regularisation & L1(1e-4)L2(1e-3) & L1(1e-4)L2(1e-3) & L1(1e-4)L2(1e-3) \\
\bottomrule
\end{tabular}
}
\end{center}
\end{table}

\begin{figure}[tbh!]
    \centering
    \includegraphics[width=0.6\linewidth]{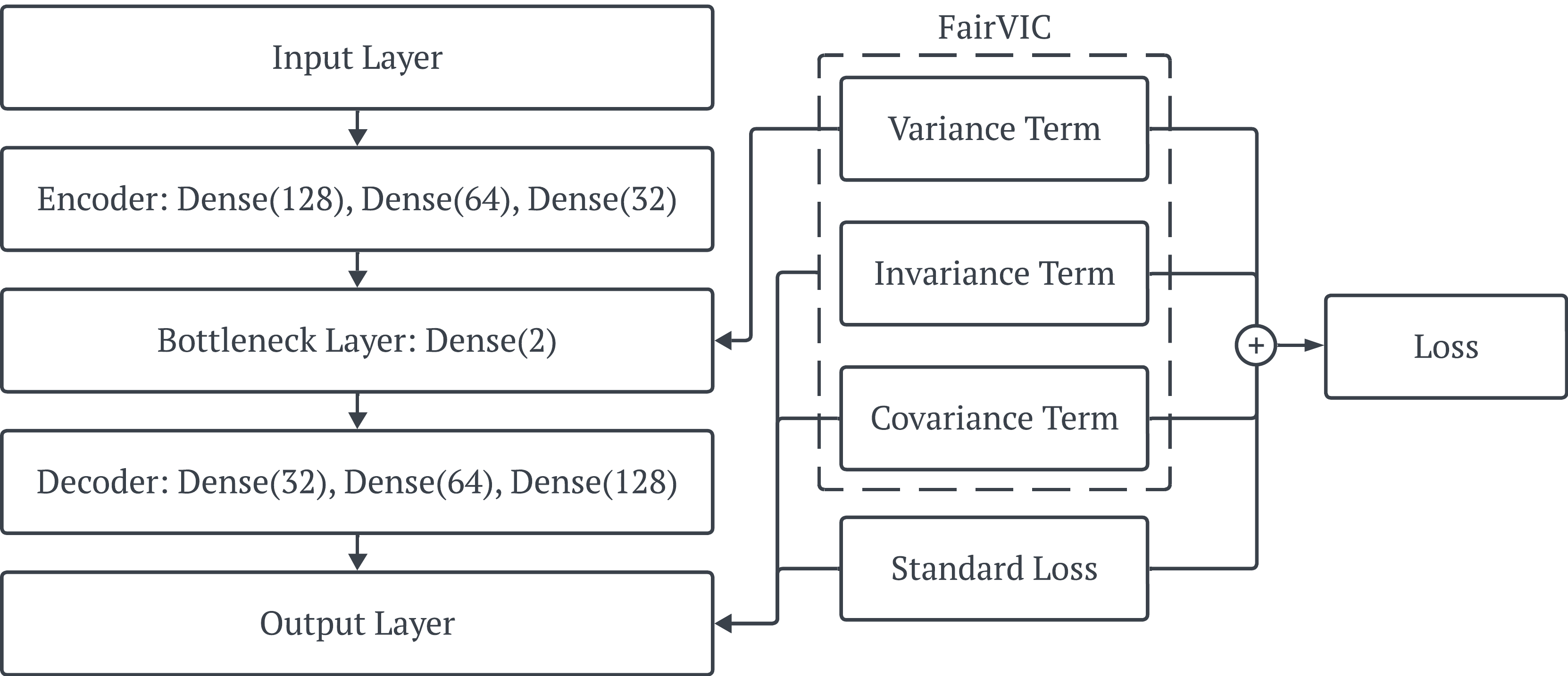}
    \caption{Network architecture for tabular data, with FairVIC loss components applied at relevant stages.}
    \label{fig:architecture}
\end{figure}

A visualisation for our neural network architecture for tabular data is seen in Figure~\ref{fig:architecture}, alongside our loss terms to illustrate where FairVIC components are applied.

All models were run with minimal and consistent data preprocessing. While some models, such as MetaFair, may underperform due to their reliance on specific sampling techniques, all comparable methods are treated uniformly as in-processing techniques. This allows them to be applied to any dataset, ensuring a fair evaluation across models.

\section{Full Training Results}
\label{sec:Full Training Results}
\begin{figure}[!htb]
\centering
    \begin{subfigure}[b]{0.9\textwidth}
    \includegraphics[width=\textwidth]{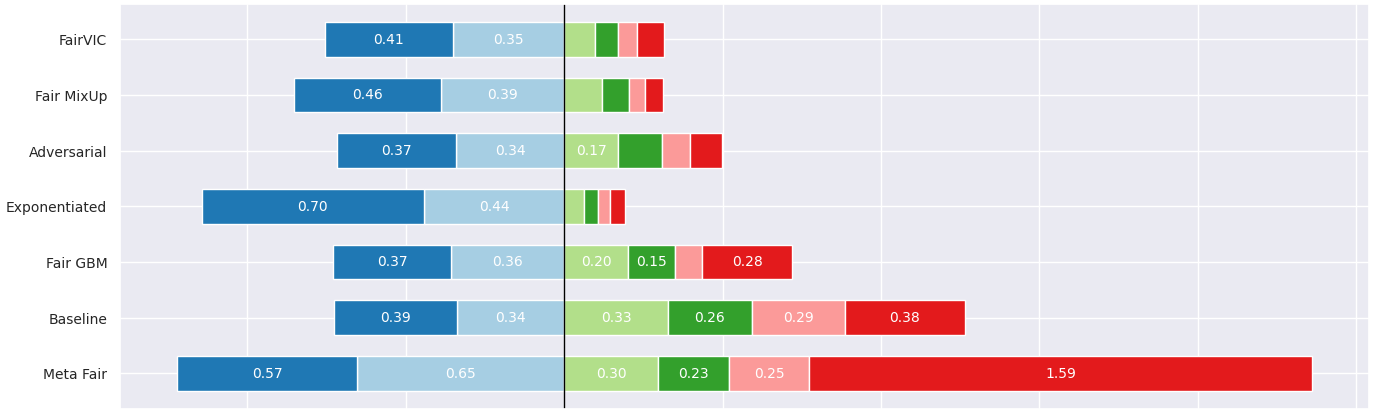}
    \caption{COMPAS dataset.}
    \label{fig:results-vis-compas}
    \end{subfigure}
    \hspace{3mm}
    \begin{subfigure}[b]{0.9\textwidth}
    \includegraphics[width=\textwidth]{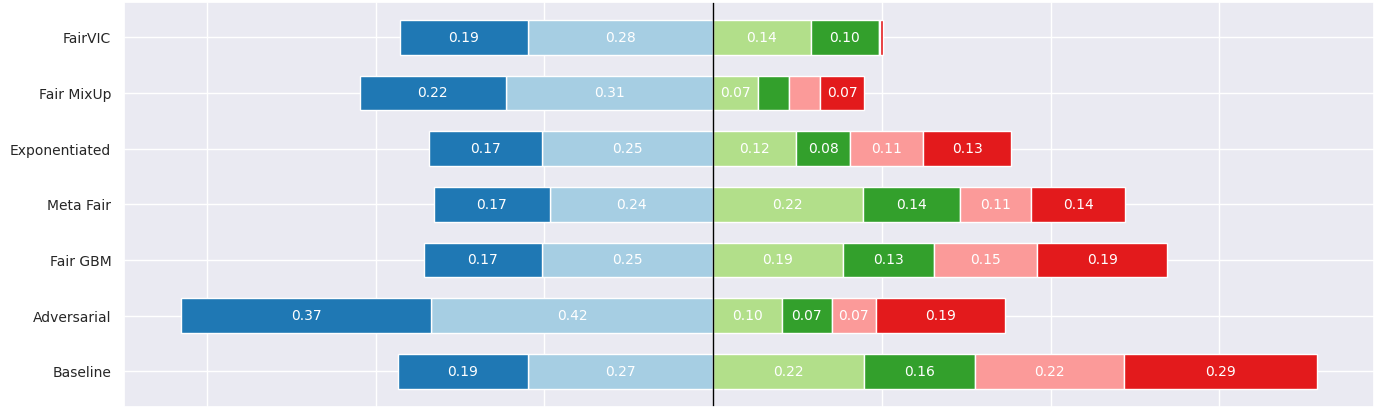}
    \caption{German Credit dataset.}
    \label{fig:results-vis-german}
    \end{subfigure}
\caption{A qualitative analysis of the absolute differences from the ideal value (e.g., perfect accuracy and fairness) in performance (left) and fairness (right) metrics of comparable techniques on the COMPAS and German Credit datasets.}
\label{fig:full-vis}
\end{figure}

In addition to the results and analysis presented in Section~\ref{sec:Evaluation}, this section provides supplementary experiments and figures. First, the qualitative visualisations for the COMPAS and German Credit can be seen in Figure~\ref{fig:full-vis}, following the analysis of the Adult Income dataset in Figure~\ref{fig:adult-vis}. Discussion on these results can be seen in Section~\ref{sec:Core Results}.

\subsection{Feature Importances}
\label{sec:Feature Importances}
\begin{figure}[tbh!]
    \includegraphics[width=\textwidth]{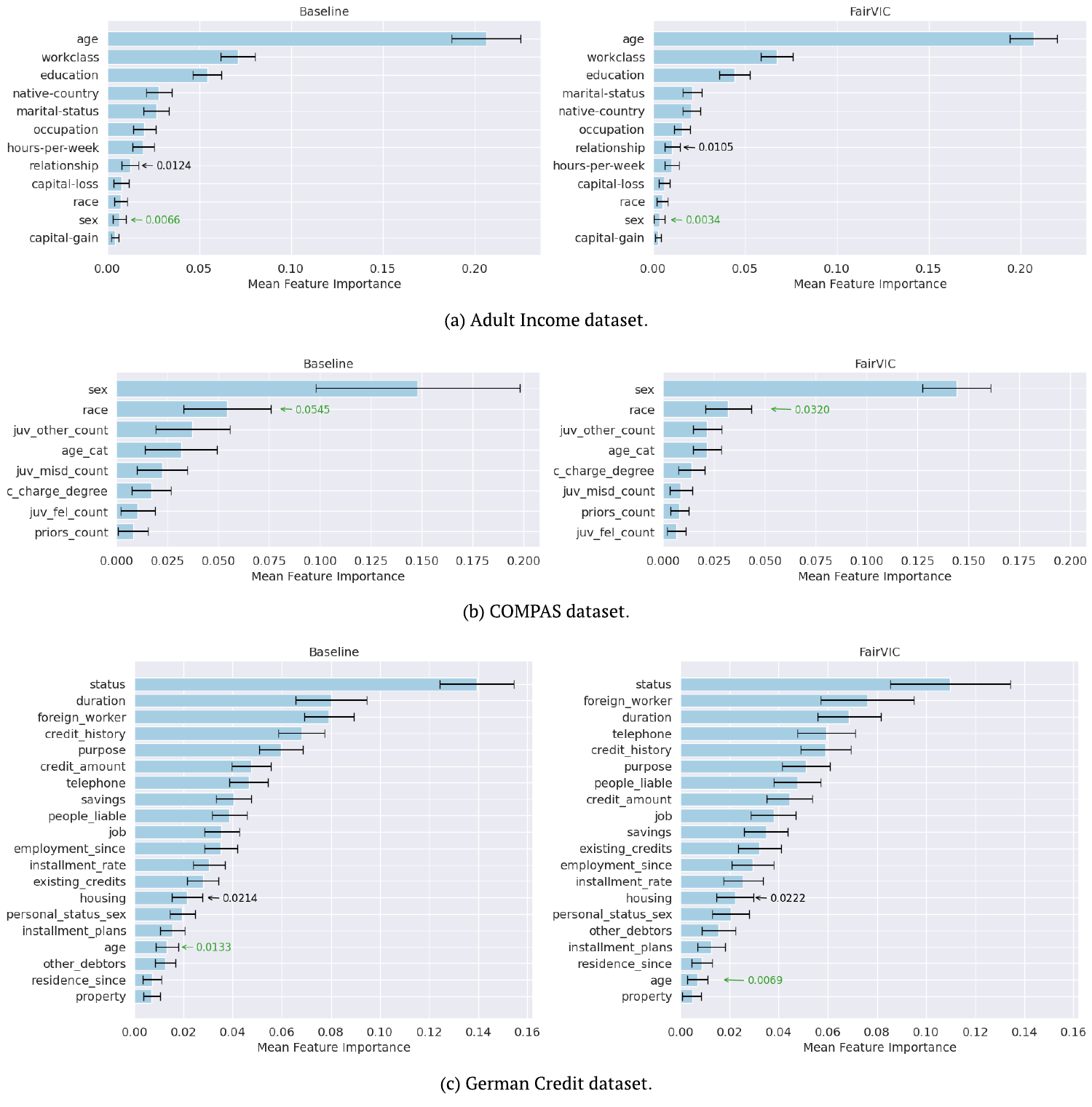}
    \caption{Mean feature importances derived from SHAP values for the baseline and FairVIC models across the three tabular datasets. The protected attribute (green) and its strong proxy variables (black) are annotated with their exact feature importance values.}
    \label{fig:full-feats-results}
\end{figure}

Figure~\ref{fig:full-feats-results} shows the feature importance of the baseline and FairVIC models across the three tabular datasets. In all baseline models, the protected attributes show some importance to the decision-making process, such as in the COMPAS dataset, where \textit{race} is a dominant feature. Combined with the results presented in Section~\ref{sec:Core Results}, this suggests that the baseline models are prone to using the protected attribute to propagate bias. Additionally, proxy variables (highlighted with their importance in black), which are strongly correlated with the protected attributes, further show how bias can be perpetuated in the baseline model. For example, in the Adult Income dataset, \textit{relationship} has a mean feature importance of $0.0124$. This indicates that even though the model appears to have limited reliance on the protected attribute \textit{sex} (which is among the least used features), it may still propagate bias through proxies such as \textit{relationship}.

In contrast, the FairVIC models for all three datasets demonstrate a strong reduction in the mean importance of protected attributes and proxy variables. This reduction is due to the three additional terms used in FairVIC- variance, invariance, and covariance. We can see that the covariance term exactly minimises the model's dependency on the protected characteristic, which, in combination with results in Section~\ref{sec:Core Results}, suggests a fairer decision-making process. The reduction in proxy variables should also be noted. Not only does FairVIC successfully reduce the reliance on the protected attribute, but it can also reduce the reliance on any features strongly correlated to the protected attribute. For example, in the Adult Income dataset, \textit{sex} and \textit{relationship} have a strong negative correlation ($-0.58$) meaning a model cannot only propagate bias through the use of \textit{sex} but also through the use of \textit{relationship} which we see the baseline model rely upon. The FairVIC model sees the mean feature importance of \textit{relationship} drop by approximately a third and the importance of \textit{sex} drop by half. This shows FairVIC’s ability to mitigate both direct and indirect biases, leading to more equitable outcomes.  On the COMPAS dataset, while \textit{race} remains as the second most important feature, its actual importance dropped by $\approx41\%$.

\subsection{Individual Fairness Results}
\label{sec:Individual Fairness Results}
Following the analysis found in Section~\ref{sec:Individual Fairness Analysis}, Table~\ref{tab:ind-results} shows the individual fairness on both the baseline and FairVIC with our recommended lambdas, and FairVIC Invariance~($\lambda_{\text{acc}}=0.1, \lambda_{\text{inv}}=0.9, \lambda_{\text{var, cov}}=0.0$) models using their absolute differences to their counterfactual model results.  In the Adult Income dataset, the mean absolute difference across all six metrics combined for the baseline model is $0.0094$, while for FairVIC invariance it is $0.0055$. In the COMPAS dataset, the mean absolute difference for the baseline model is $0.0285$, while for FairVIC Invariance it is $0.0050$. Finally, for the German Credit dataset the mean absolute difference for the baseline model is $0.0277$, while for FairVIC Invariance it is $0.0108$. FairVIC's invariance term, designed to enhance individual fairness, proves to be effective. The FairVIC invariance model consistently achieves significantly absolute differences, demonstrating the success of the approach. In our selection of FairVIC terms, we prioritize group fairness by weighting invariance lower, yet the model still maintains low counterfactual absolute differences.

For discussion on the FairVIC Invariance model individual fairness results, see Section~\ref{sec:Individual Fairness Analysis}.

\begin{table}[tbh!]
\caption{Counterfactual (CF) model results and absolute differences (ADs) for the baseline, FairVIC ($\lambda_{\text{acc, var, inv}}=0.1, \lambda_{\text{cov}}=0.7$), and FairVIC Invariance~($\lambda_{\text{acc}}=0.1, \lambda_{\text{inv}}=0.9, \lambda_{\text{var, cov}}=0.0$) models.}
\label{tab:ind-results}
\resizebox{0.9\textwidth}{!}{%
\begin{tabular}{clllcccc}
\toprule
\multicolumn{2}{c}{} &
\multicolumn{2}{c}{Performance Metrics} &
\multicolumn{4}{c}{Fairness Metrics} \\
\cmidrule(lr){3-4} \cmidrule(lr){5-8}
\multicolumn{1}{c}{Dataset} &
\multicolumn{1}{c}{Model} &
\multicolumn{1}{c}{Accuracy} &
\multicolumn{1}{c}{F1 Score} &
\multicolumn{1}{c}{Equalized Odds} &
\multicolumn{1}{c}{Absolute Odds} &
\multicolumn{1}{c}{Statistical Parity} &
\multicolumn{1}{c}{Disparate Impact} \\
\midrule
 &
  Baseline  &
   0.8444 $\pm$ 0.0065&
   0.6685 $\pm$ 0.0118&
   0.1330 $\pm$ 0.0317&
   0.1172 $\pm$ 0.0289&
   -0.2173 $\pm$ 0.0291&
   0.2853 $\pm$ 0.0329\\  
 &
  Baseline CF  &
   0.8444 $\pm$ 0.0059&
   0.6649 $\pm$ 0.0114&
   0.1208 $\pm$ 0.0286&
   0.1026 $\pm$ 0.0285&
   -0.2069 $\pm$ 0.0290&
   0.3006 $\pm$ 0.0347\\  
 &
Baseline AD &
  0.0000 $\pm$ 0.0032&
  0.0036 $\pm$ 0.0089&
  0.0123 $\pm$ 0.0329&
  0.0147 $\pm$ 0.0211&
  0.0104 $\pm$ 0.0147&
  0.0152 $\pm$ 0.0193\\
\cmidrule(lr){2-8}
 &
  FairVIC Invariance  &
   0.8150 $\pm$ 0.0053&
   0.4281 $\pm$ 0.0938&
   0.0437 $\pm$ 0.0363&
   0.0342 $\pm$ 0.0330&
   -0.0811 $\pm$ 0.0516&
   0.3199 $\pm$ 0.0383\\ 
 &
 FairVIC Invariance CF  &
   0.8147 $\pm$ 0.0067&
   0.4242 $\pm$ 0.0954&
   0.0438 $\pm$ 0.0332&
   0.0303 $\pm$ 0.0286&
   -0.0752 $\pm$ 0.0479&
   0.3388 $\pm$ 0.0541\\  
 &
  FairVIC Invariance AD &
  0.0003 $\pm$ 0.0039&
  0.0039 $\pm$ 0.0472&
  0.0002 $\pm$ 0.0169&
  0.0040 $\pm$ 0.0145&
  0.0059 $\pm$ 0.0228&
  0.0189 $\pm$ 0.0445\\ 
\cmidrule(lr){2-8}
 &
  FairVIC  &
  0.8284 $\pm$ 0.0088&
  0.5314 $\pm$ 0.0509&
  0.2993 $\pm$ 0.0683&
  0.1637 $\pm$ 0.0371&
  -0.0088 $\pm$ 0.0249&
  0.9803 $\pm$ 0.2220\\  
&
  FairVIC CF &
  0.8310 $\pm$ 0.0075&
   0.5430 $\pm$ 0.0382&
  0.2793 $\pm$ 0.0563&
  0.1524 $\pm$ 0.0326&
  -0.0166 $\pm$ 0.0235&
  0.9015 $\pm$ 0.1450\\ 
 \multirow{-9}{*}{\cellcolor{white}\makecell{Adult\\Income}} &
  FairVIC AD &
  0.0007 $\pm$ 0.0055&
  0.0243 $\pm$ 0.0368&
  0.0240 $\pm$ 0.0397&
  0.0152 $\pm$ 0.0243&
  0.0022 $\pm$ 0.0130&
  0.0313 $\pm$ 0.1266\\
\midrule
 &
  Baseline  &
   0.6622 $\pm$ 0.0150&
   0.6118 $\pm$ 0.0252&
   0.3281 $\pm$ 0.0574&
   0.2635 $\pm$ 0.0452&
   -0.2941 $\pm$ 0.0459&
   0.6223 $\pm$ 0.0504\\  
 &
  Baseline CF  &
   0.6651 $\pm$ 0.0183&
   0.6285 $\pm$ 0.0389&
   0.2707 $\pm$ 0.0599&
   0.2237 $\pm$ 0.0608&
   -0.2588 $\pm$ 0.0585&
   0.6415 $\pm$ 0.0763\\ 
 &
 Baseline AD &
  0.0028 $\pm$ 0.0054&
  0.0167 $\pm$ 0.0190&
  0.0575 $\pm$ 0.0516&
  0.0398 $\pm$ 0.0334&
  0.0353 $\pm$ 0.0309&
  0.0192 $\pm$ 0.0329\\ 
\cmidrule(lr){2-8}
 &
FairVIC Invariance  &
   0.6571 $\pm$ 0.0121&
   0.6232 $\pm$ 0.0384&
   0.2618 $\pm$ 0.0412&
   0.2101 $\pm$ 0.0266&
   -0.2435 $\pm$ 0.0264&
   0.6530 $\pm$ 0.0551\\ 
 &
  FairVIC Invariance CF  &
   0.6564 $\pm$ 0.0109&
   0.6130 $\pm$ 0.0366&
   0.2689 $\pm$ 0.0341&
   0.2117 $\pm$ 0.0333&
   -0.2438 $\pm$ 0.0324&
   0.6633 $\pm$ 0.0642\\  
 &
  FairVIC Invariance AD &
  0.0007 $\pm$ 0.0072&
  0.0102 $\pm$ 0.0232&
  0.0071 $\pm$ 0.0332&
  0.0016 $\pm$ 0.0166&
  0.0003 $\pm$ 0.0159&
  0.0103 $\pm$ 0.0256\\
\cmidrule(lr){2-8}
 &
  FairVIC  &
  0.6501 $\pm$ 0.0173 &
  0.5934 $\pm$ 0.0357 &
  0.0976 $\pm$ 0.0375 &
  0.0719 $\pm$ 0.0305 &
  -0.0602 $\pm$ 0.0678 &
  0.9139 $\pm$ 0.1135 \\  
 &
  FairVIC CF  &
  0.6295 $\pm$ 0.0392 &
  0.5154 $\pm$ 0.1767 &
  0.0771 $\pm$ 0.0532 &
  0.0506 $\pm$ 0.0353 &
  -0.0394 $\pm$ 0.0609 &
  0.9489 $\pm$ 0.1057 \\ 
 \multirow{-9}{*}{\cellcolor{white}COMPAS} &
  FairVIC AD &
  0.0205 $\pm$ 0.0419&
  0.0780 $\pm$ 0.1874&
  0.0204 $\pm$ 0.0368&
  0.0213 $\pm$ 0.0336&
  0.0209 $\pm$ 0.0450&
  0.0350 $\pm$ 0.0723\\
\midrule
 &
  Baseline  &
   0.7255 $\pm$ 0.0284 &
   0.8077 $\pm$ 0.0275 &
   0.2234 $\pm$ 0.0974 &
   0.1641 $\pm$ 0.0936 &
   -0.2218 $\pm$ 0.0901 &
   0.7140 $\pm$ 0.1203 \\ 
 &
  Baseline CF  &
   0.7010 $\pm$ 0.0371&
   0.7889 $\pm$ 0.0388&
   0.2222 $\pm$ 0.0979&
   0.1677 $\pm$ 0.0830&
   -0.1678 $\pm$ 0.1171&
   0.7782 $\pm$ 0.1495\\ 
 &
 Baseline AD &
  0.0245 $\pm$ 0.0294&
  0.0189 $\pm$ 0.0257&
  0.0012 $\pm$ 0.0576&
  0.0036 $\pm$ 0.0454&
  0.0540 $\pm$ 0.0520&
  0.0641 $\pm$ 0.0700\\
\cmidrule(lr){2-8}
 &
  FairVIC Invariance  &
   0.7165 $\pm$ 0.0356&
   0.7917 $\pm$ 0.0319&
   0.1367 $\pm$ 0.0798&
   0.0964 $\pm$ 0.0521&
   -0.0600 $\pm$ 0.1090&
   0.9113 $\pm$ 0.1625\\ 
 &
  FairVIC Invariance CF  &
   0.7250 $\pm$ 0.0356&
   0.7961 $\pm$ 0.0412&
   0.1257 $\pm$ 0.0793&
   0.0927 $\pm$ 0.0603&
   -0.0759 $\pm$ 0.0724&
   0.8902 $\pm$ 0.0972\\  
 &
  FairVIC Invariance AD &
  0.0085 $\pm$ 0.0272&
  0.0044 $\pm$ 0.0316&
  0.0109 $\pm$ 0.0843&
  0.0036 $\pm$ 0.0548&
  0.0159 $\pm$ 0.0689&
  0.0211 $\pm$ 0.0856\\
\cmidrule(lr){2-8}
 &
  FairVIC  &
  0.7250 $\pm$ 0.0239 &
  0.8108 $\pm$ 0.0237 &
  0.1443 $\pm$ 0.0796 &
  0.1017 $\pm$ 0.0464 &
  0.0016 $\pm$ 0.0604 &
  1.0037 $\pm$ 0.0764 \\   
 &
  FairVIC CF  &
  0.7380 $\pm$ 0.0223 &
  0.8248 $\pm$ 0.0134 &
  0.1466 $\pm$ 0.0943 &
  0.1002 $\pm$ 0.0507 &
  -0.0017 $\pm$ 0.0768 &
  1.0000 $\pm$ 0.0979 \\ 
 \multirow{-9}{*}{\cellcolor{white}\makecell{German\\Credit}} &
  FairVIC AD &
  0.0130 $\pm$ 0.0198&
  0.0140 $\pm$ 0.0239&
  0.0023 $\pm$ 0.0292&
  0.0014 $\pm$ 0.0181&
  0.0033 $\pm$ 0.0584&
  0.0036 $\pm$ 0.0759\\
\bottomrule
\end{tabular}
}
\end{table}

\subsection{Hyperparameter Recommendations}
\label{sec:Hyperparameter Recommendations}
The weights for the loss terms in FairVIC (\(\lambda_{\text{acc}}, \lambda_{\text{var}}, \lambda_{\text{inv}}, \lambda_{\text{cov}}\)) were chosen based on insights from our ablation studies. We have outlined the weights we used in our evaluation in Table~\ref{tab:Final Lambdas}.

\begin{table}
    \centering
    \caption{Our final lambda selection and recommendations for each of the seven datasets evaluated upon.}
    \label{tab:Final Lambdas}
    \resizebox{0.9\textwidth}{!}{%
    \begin{tabular}{cccccccc} \toprule 
         &  Adult Income & Compas & German Credit & CivilComments- WILDS & BiasBios & CelebA & UTKFace \\ \midrule
         $\lambda_{\text{acc}}$&  0.20&  0.10&  0.10&  0.10&  0.05&  0.10& 0.20\\
         $\lambda_{\text{var}}$&  0.10&  0.10&  0.10&  0.10&  0.05&  0.05& 0.10\\
         $\lambda_{\text{inv}}$&  0.10&  0.10&  0.10&  0.10&  0.05&  0.05& 0.10\\
         $\lambda_{\text{cov}}$&  0.60&  0.70&  0.70&  0.70&  0.85&  0.80& 0.60\\ \bottomrule
    \end{tabular}
    }
\end{table}


To help users configure FairVIC effectively, we provide a recommended starting point and guidance for adapting the loss term weights to suit different fairness and performance goals. While our paper focuses on generalisable defaults, in a real-world deployment, one could envisage a tuning process to identify optimal results for a given application or domain.

We recommend starting with the configuration $\lambda_{\text{acc}} = 0.1$, $\lambda_{\text{var}} = 0.1$, $\lambda_{\text{inv}} = 0.1$, and $\lambda_{\text{cov}} = 0.7$, which provided strong results across multiple datasets including COMPAS, German Credit, and CivilComments-WILDS. This balanced setting encourages individual fairness, supports diverse representations, and strongly targets group fairness without significantly compromising accuracy. The decision to use a relatively low weight for accuracy (\(\lambda_{\text{acc}} = 0.1\)) stems from the equal ablation study results, which demonstrated that this value achieves the best fairness-accuracy trade-off for these datasets. Group fairness is given significant emphasis, as shown by the higher weight assigned to the covariance term (\(\lambda_{\text{cov}} = 0.7\)), which plays a key role in minimizing disparities across protected groups. Meanwhile, the variance (\(\lambda_{\text{var}}\)) and invariance (\(\lambda_{\text{inv}}\)) terms were assigned a weight of 0.1, as this value still allowed for their individual fairness aims to be achieved effectively, thus balancing all fairness and accuracy objectives.

To adjust for different dataset characteristics or application goals, users should consider the intended role of each loss term:

If accuracy needs improvement while fairness is already strong, increasing the accuracy term weight can help the model prioritise predictive performance. For instance, in the Adult Income and UTKFace datasets, increasing $\lambda_{\text{acc}}$ to 0.2 (while slightly lowering $\lambda_{\text{cov}}$) led to better trade-offs. This is likely because both datasets have relatively high-quality features and well-separated class distributions, allowing the model to benefit from a greater emphasis on discriminative capacity once basic fairness has been addressed.

When group fairness is a higher priority, the covariance term should be weighted more heavily. This term explicitly penalises statistical dependence between predictions and group membership, and is most impactful when datasets show strong baseline disparities between protected and privileged groups. For example, image-based datasets such as CelebA often contain visually encoded group cues that strongly correlate with target labels, while text-based datasets like BiasBios may exhibit linguistic patterns linked to social or demographic attributes. In such cases, increasing $\lambda_{\text{cov}}$ (e.g. to 0.8 or 0.85) helps reduce spurious correlations and promotes more group-independent predictions.

To enhance individual fairness, a higher weight on the invariance term helps the model maintain consistency under counterfactual changes to the protected attribute. This is especially useful for datasets where protected features are not strongly entangled with the label, such that counterfactual consistency is a plausible fairness goal. Although our default already includes invariance, users with specific fairness requirements may find that increasing $\lambda_{\text{inv}}$ offers more robust guarantees.

In practice, the ideal configuration depends on the dataset size, complexity, and the fairness-performance trade-off required by the application. Larger or more complex datasets may require a slightly higher weight on the accuracy term to avoid underfitting, while fairness-sensitive domains may justify higher weights on the fairness terms.

\begin{itemize}
    \item \textbf{Default configuration:} \(\lambda_{\text{acc}} = 0.1,\ \lambda_{\text{var}} = 0.1,\ \lambda_{\text{inv}} = 0.1,\ \lambda_{\text{cov}} = 0.7\)
    \item \textbf{To improve accuracy:} Increase \(\lambda_{\text{acc}}\)
    \item \textbf{To emphasise group fairness:} Increase \(\lambda_{\text{cov}}\)
    \item \textbf{To emphasise individual fairness:} Increase \(\lambda_{\text{inv}}\)
\end{itemize}

If users have a specific optimisation goal—such as maximising a particular fairness or accuracy metric—a targeted grid search over the loss weights may be appropriate, though this falls outside the primary scope of our work.

\subsection{Model Representation Analysis}
\label{sec:Model Representation Analysis}
\begin{figure}[tbh!]
    \centering
    \includegraphics[width=0.85\linewidth]{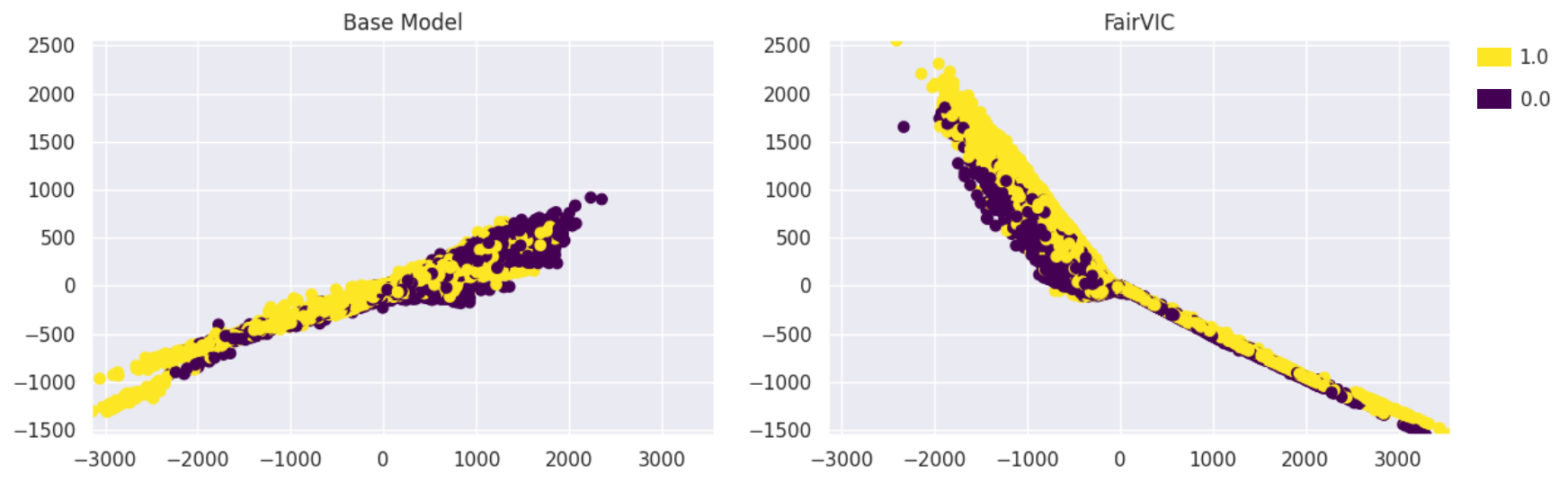}
    \caption{An example latent space visualization from one random seed of a baseline model and a FairVIC model on the Adult Income dataset. Subgroup (1) represents male individuals, and subgroup (0) represents female individuals.}
    \label{fig:latentspace}
\end{figure}

An example latent space visualisation from the baseline model and FairVIC can be seen in Figure~\ref{fig:latentspace}. In the baseline model, we observe a separation between subgroups, where women (subgroup 0) are predominantly located in the upper region and men (subgroup 1) in the lower region of the latent space. This separation suggests that the baseline model's representations may be influenced by the protected attribute, leading to the biased decision-making reported in Table~\ref{tab:CoreResults}. In contrast, the FairVIC model shows a more condensed and overlapping distribution of both subgroups within the same latent space. This, alongside results in Table~\ref{tab:CoreResults} and feature importance in Figure~\ref{fig:full-feats-results}, indicates that FairVIC has successfully reduced the model’s reliance on the protected characteristic and any proxy variables, thereby promoting more equitable representations. The overlapping and compact structure in the FairVIC latent space demonstrates that similar data points, regardless of their subgroup membership, are mapped closer together, ensuring that the model’s predictions are not unfairly biased towards one group over the other.

\subsection{Model Optimization Analysis}
\label{sec:Model Optimization Analysis}
\begin{figure}[htb]
    \centering
    \includegraphics[width=0.8\linewidth]{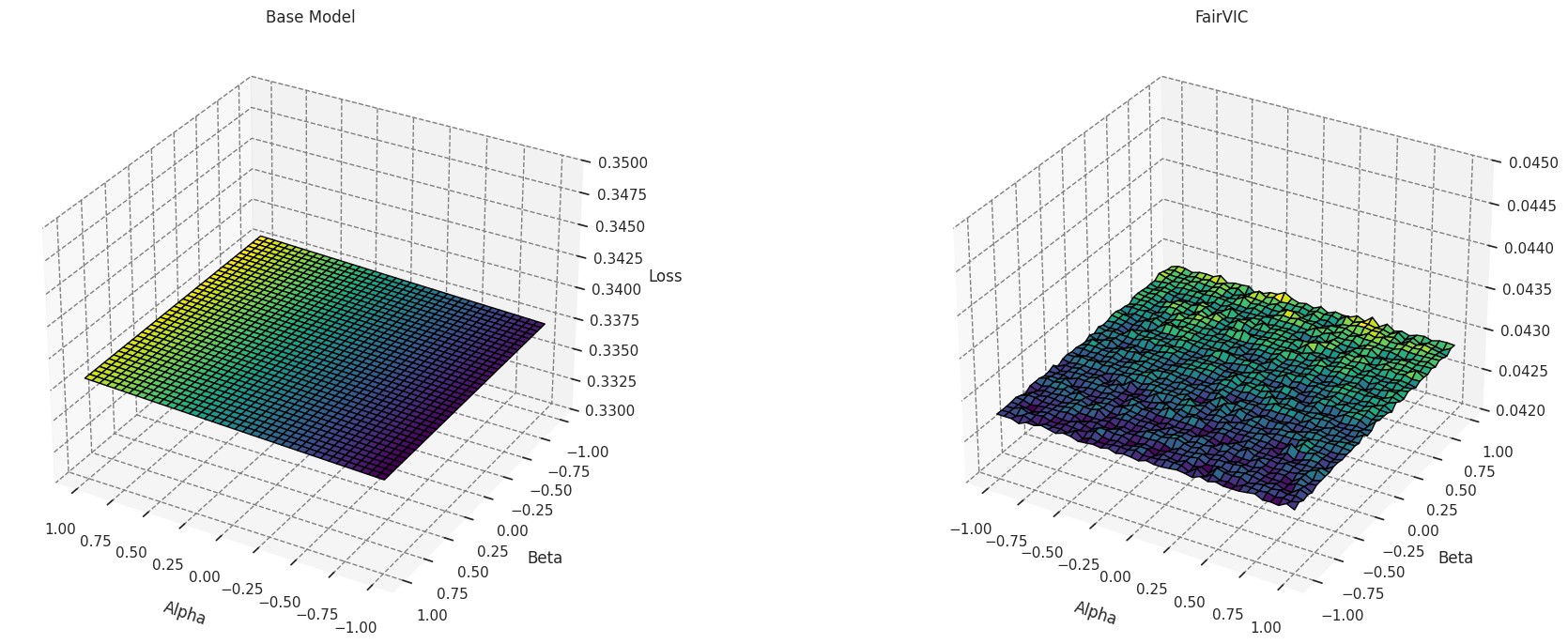}
    \caption{An example loss landscape visualisation from one random seed of a baseline model and a FairVIC model on the Adult Income dataset.}
    \label{fig:loss-landscape}
\end{figure}

Figure~\ref{fig:loss-landscape} illustrates the loss landscapes of the baseline and FairVIC models on the Adult Income dataset. Both models exhibit smooth loss surfaces, indicating that they are relatively well-optimized. The baseline model (left) shows a stable loss landscape with a slight gradient. The FairVIC model (right), despite incorporating additional fairness constraints, maintains a similarly smooth surface albeit with tiny peaks in various places. This demonstrates that the inclusion of variance, invariance, and covariance terms in the loss function does not introduce instability or optimisation challenges.

\FloatBarrier

\subsection{Theoretical Analysis and Discussion}
\label{Theoretical Analysis and Discussion}

In this section, we theoretically analyse FairVIC and show how each individual loss term is sub-differentiable.

\subsubsection{Theorem~\ref{thm:1}}

\begin{theorem}
Each individual term in FairVIC $L_{var}, L_{inv}, L_{cov}$ is sub-differentiable everywhere in the model's parameters $\theta$.
\label{thm:1}
\end{theorem}

\begin{proof}
The variance term is defined as:

\begin{equation}
L_{\text{var}} = \frac{1}{d} \sum_{j=1}^{d} \max(0, \gamma - \sigma_j(z))
\end{equation}

where $z$ is the latent embeddings for the input $x$. $z = g_{\theta}(x)$ is continuous in $\theta$, where $g_{\theta}$ is the function/layer that maps input $x$ to the latent embedding $z$. $\sigma(z)$ is the standard deviation of a continuous variable in a finite sample, which is continuous except at rare instances where all $z_j$ are identical. Even in this degenerate case, $\sigma(\cdot)$ is sub-differentiable. The $\text{max}(0, \cdot)$ operator is only non-differentiable at $0$, where the sub-derivative set is $[0,1]$. Hence $\text{max}(0, \cdot)$ is sub-differentiable w.r.t $\theta$. The invariance term is defined as:

\begin{equation}
L_{\text{inv}} = \frac{1}{N} \sum_{i=1}^{N} \left( \hat{y}_i - \hat{y}_i^{*} \right)^2
\end{equation}

where $\hat{y}_i$ is the model's predictions, and $\hat{y}_i^{*}$ is the model's predictions where the protected attribute is flipped. As $\hat{y}_i$ is differentiable in $\theta$, then $\left( \hat{y}_i - \hat{y}_i^{*} \right)^2$ is differentiable as it is the composition of smooth functions. The covariance term is defined as:

\begin{equation}
    \new{L_{\text{cov}} = \frac{1}{N} \sqrt{\sum_{i=1}^{N} \left( \left(\hat{y}_i - \mathbb{E}[\hat{y}]\right) \cdot P_i \right)^2}}
\end{equation}

where $\sum_{i=1}^{N}\left(\left(\hat{y} - \mathbb{E}[\hat{y}]\right) \cdot P\right)^2$ is a sum of squares, which is smooth and differentiable. The square root is differentiable for non-zero input and sub-differentiable at $0$.

Each of the three terms is (sub-)differentiable everywhere in $\theta$. Hence a gradient-based or subgradient-based method can be applied directly with FairVIC.

\end{proof}

\subsection{Trade-Off Discussion}
\label{sec:Trade-Off Discussion}
The design of FairVIC is grounded in the recognition that group fairness metrics such as Statistical Parity Difference, Equalized Odds Difference, and Disparate Impact capture distinct and often conflicting notions of fairness. Foundational theoretical work by \citet{kleinberg2016inherent} has shown that, under realistic conditions- such as differing base rates between groups- it is impossible for any prediction system to simultaneously satisfy multiple fairness criteria, such as calibration and equalized odds, without introducing trade-offs.

FairVIC does not attempt to satisfy these fairness criteria simultaneously in a formal sense. Rather than optimising for any specific fairness definition directly, FairVIC introduces inductive biases at the representation level through variance, invariance, and covariance regularisation. These objectives are agnostic to downstream fairness metrics and aim to induce feature representations that are disentangled from protected attributes while preserving predictive signal.

The improvements observed across multiple fairness metrics are therefore not the result of directly encoding incompatible fairness constraints, but an emergent property of the learned representations. As shown in Table~\ref{tab:CoreResults}, FairVIC achieves significant reductions in Statistical Parity Difference and Equalized Odds Difference across all datasets, and improves Disparate Impact in most cases. In the case of our experiments, these fairness improvements come with modest reductions in predictive performance- consistent with expected trade-offs. Sometimes, improvements to one fairness metric  such as Disparate Impact- slightly affect the results for another fairness metric, highlighting that such trade-offs are context-dependent and not inevitable.

Importantly, FairVIC does not commit to a normative fairness definition during training. This design choice reflects a practical and ethical consideration: in many real-world settings, there may be no consensus on which fairness metric best captures the relevant notion of harm or justice. By focusing on representation-level properties rather than metric-specific constraints, FairVIC supports post hoc evaluation across multiple fairness metrics, encouraging empirical pluralism and transparency in fairness assessments.

In summary, FairVIC does not circumvent impossibility theorems, nor does it attempt to. Rather, it sidesteps the need to encode conflicting fairness notions directly, and instead fosters representation learning that generalises well across subgroups. The observed metric improvements - and their trade-offs - are empirical outcomes of this strategy, not theoretical contradictions.

\section{Lambda Ablation Study Results}
\label{sec:Lambda Ablation Study Results}
Tables~\ref{tab:equal_vic_results} and~\ref{tab:individual_vic_results} show the full results for each model when the weights on the FairVIC terms are adapted. Table~\ref{tab:equal_vic_results} shows the effect of changing $\lambda_{\text{acc}}$ while keeping the FairVIC terms equal so that $\lambda_{\text{var, inv, cov}}=\frac{1-\lambda_{\text{acc}}}{3}$, where $0<\lambda_{\text{acc}}<1$, and Table~\ref{tab:individual_vic_results} sets $\lambda_{\text{acc}}=0.1$, and suppresses one or two FairVIC terms to explore the effect of only utilising one or two term(s) at a time. For full discussion and analysis of the results of the lambda ablation study, see Section~\ref{sec:Lambda Ablation Study Analysis}.

\begin{table}[tb!]
\caption{Performance and fairness results for FairVIC on the three tabular datasets, where the FairVIC terms are weighted equally, such that $\lambda_{\text{acc}} + \lambda_{\text{var}} + \lambda_{\text{inv}} + \lambda_{\text{cov}} = 1$.}
\label{tab:equal_vic_results}
\resizebox{0.9\textwidth}{!}{%
\begin{tabular}{ccccccccc}
\toprule
\multicolumn{3}{c}{} &
\multicolumn{2}{c}{Performance Metrics} &
\multicolumn{4}{c}{Fairness Metrics} \\
\cmidrule(lr){4-5} \cmidrule(lr){6-9}
\multicolumn{1}{c}{Dataset} &
\multicolumn{1}{c}{$\lambda_{\text{acc}}$} &
\multicolumn{1}{c}{$\lambda_{\text{var,inv,cov}}$} &
\multicolumn{1}{c}{Accuracy} &
\multicolumn{1}{c}{F1 Score} &
\multicolumn{1}{c}{Equalized Odds} &
\multicolumn{1}{c}{Absolute Odds} &
\multicolumn{1}{c}{Statistical Parity} &
\multicolumn{1}{c}{Disparate Impact} \\
\midrule
 &
  0.10 &
  0.30 &
   0.8157 $\pm$ 0.0061&
   0.4622 $\pm$ 0.0412&
   0.3506 $\pm$ 0.0560&
   0.1929 $\pm$ 0.0328&
   0.0191 $\pm$ 0.0222&
   1.2542 $\pm$ 0.2653\\ 
 &
  0.20 &
  $0.2\overline{6}$ &
   0.8358 $\pm$ 0.0084&
   0.5500 $\pm$ 0.0463&
   0.2321 $\pm$ 0.0784&
   0.1226 $\pm$ 0.0442&
   -0.0339 $\pm$ 0.0302&
   0.7970 $\pm$ 0.2118\\ 
 &
  0.30 &
  $0.2\overline{3}$ &
   0.8448 $\pm$ 0.0053&
   0.6061 $\pm$ 0.0421&
   0.0918 $\pm$ 0.0507&
   0.0550 $\pm$ 0.0229&
   -0.0983 $\pm$ 0.0341&
   0.5073 $\pm$ 0.0952\\ 
 &
  0.40 &
  0.20 &
   0.8481 $\pm$ 0.0033&
   0.6354 $\pm$ 0.0253&
   0.0560 $\pm$ 0.0137&
   0.0433 $\pm$ 0.0102&
   -0.1339 $\pm$ 0.0276&
   0.4069 $\pm$ 0.0434\\ 
\multirow{-5}{*}{\makecell{Adult\\Income}} &
  0.50 &
  $0.1\overline{6}$ &
   0.8506 $\pm$ 0.0052&
   0.6564 $\pm$ 0.0110&
   0.0630 $\pm$ 0.0117&
   0.0473 $\pm$ 0.0175&
   -0.1561 $\pm$ 0.0161&
   0.3686 $\pm$ 0.0301\\
\midrule
 &
  0.10 &
  0.30 &
   0.6618 $\pm$ 0.0130&
   0.6061 $\pm$ 0.0197&
   0.1881 $\pm$ 0.0412&
   0.1451 $\pm$ 0.0317&
   -0.1754 $\pm$ 0.0324&
   0.7533 $\pm$ 0.0442\\ 
 &
  0.20 &
  $0.2\overline{6}$ &
   0.6661 $\pm$ 0.0114&
   0.6661 $\pm$ 0.0114&
   0.2000 $\pm$ 0.0466&
   0.1448 $\pm$ 0.0339&
   -0.1805 $\pm$ 0.0306&
   0.7391 $\pm$ 0.0344\\ 
 &
  0.30 &
  $0.2\overline{3}$ &
   0.6606 $\pm$ 0.0091&
   0.6162 $\pm$ 0.0266&
   0.1754 $\pm$ 0.0615&
   0.1326 $\pm$ 0.0485&
   -0.1687 $\pm$ 0.0426&
   0.7545 $\pm$ 0.0564\\ 
 &
  0.40 &
  0.20 &
   0.6643 $\pm$ 0.0094&
   0.6162 $\pm$ 0.0202&
   0.1946 $\pm$ 0.0400&
   0.1433 $\pm$ 0.0345&
   -0.1797 $\pm$ 0.0313&
   0.7457 $\pm$ 0.0363\\ 
\multirow{-5}{*}{COMPAS} &
    0.50 &
    $0.1\overline{6}$ &
   0.6681 $\pm$ 0.0142&
   0.6239 $\pm$ 0.0192&
   0.2037 $\pm$ 0.0500&
   0.1654 $\pm$ 0.0527&
   -0.1988 $\pm$ 0.0526&
   0.7221 $\pm$ 0.0620\\
\midrule
 &
  0.10 &
  0.30 &
   0.7160 $\pm$ 0.0431&
   0.8059 $\pm$ 0.0351&
   0.1034 $\pm$ 0.0429&
   0.0704 $\pm$ 0.0300&
   -0.0298 $\pm$ 0.0612&
   0.9574 $\pm$ 0.0842\\
   &
  0.20 &
  $0.2\overline{6}$ &
   -0.0298 $\pm$ 0.0612&
   0.8112 $\pm$ 0.0239&
   0.1305 $\pm$ 0.0754&
   0.0915 $\pm$ 0.0506&
   -0.0190 $\pm$ 0.0970&
   0.9791 $\pm$ 0.1282\\  &
  0.30 &
  $0.2\overline{3}$ &
   0.7205 $\pm$ 0.0286&
   0.8042 $\pm$ 0.0229&
   0.1189 $\pm$ 0.0593&
   0.0864 $\pm$ 0.0522&
   -0.0545 $\pm$ 0.0842&
   0.9305 $\pm$ 0.1154\\ 
 &
  0.40 &
  0.20 &
   0.7265 $\pm$ 0.0270&
   0.8096 $\pm$ 0.0226&
   0.1222 $\pm$ 0.1240&
   0.0815 $\pm$ 0.0783&
   -0.0767 $\pm$ 0.0880&
   0.9004 $\pm$ 0.1189\\
\multirow{-5}{*}{\makecell{German\\Credit}} &
  0.50 &
  $0.1\overline{6}$ &
   0.7175 $\pm$ 0.0211&
   0.8029 $\pm$ 0.0199&
   0.1073 $\pm$ 0.0549&
   0.0745 $\pm$ 0.0406&
   -0.0851 $\pm$ 0.0605&
   0.8866 $\pm$ 0.0839\\
\bottomrule
\end{tabular}
}
\end{table}

\begin{figure}[tb!]
    \centering
    \includegraphics[width=0.95\linewidth]{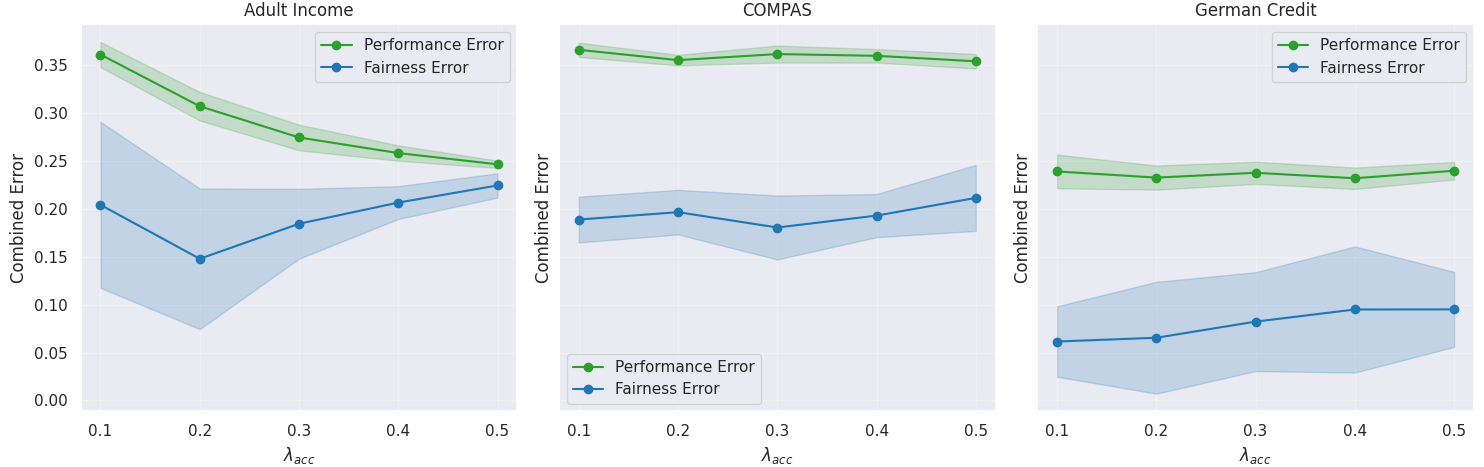}
    \caption{Absolute difference from the ideal value for performance (green) and fairness (blue) metrics of FairVIC with varying $\lambda_{\text{acc}}$ values across all tabular datasets. The FairVIC terms are weighted equally, such that $\lambda_{\text{acc}} + \lambda_{\text{var}} + \lambda_{\text{inv}} + \lambda_{\text{cov}} = 1$.}
    \label{fig:trade-off-graph}
\end{figure}

\begin{table}[tb!]
\caption{Performance and fairness results for FairVIC on the three tabular datasets, where only one or two FairVIC terms ($\lambda_{\text{var}}$, $\lambda_{\text{inv}}$, or $\lambda_{\text{cov}}$) are weighted at a time.}
\label{tab:individual_vic_results}
\resizebox{0.9\textwidth}{!}{%
\begin{tabular}{ccccccccccc}
\toprule
\multicolumn{5}{c}{} &
\multicolumn{2}{c}{Performance Metrics} &
\multicolumn{4}{c}{Fairness Metrics} \\
\cmidrule(lr){6-7} \cmidrule(lr){8-11}
\multicolumn{1}{c}{Dataset} &
\multicolumn{1}{c}{$\lambda_{\text{acc}}$} &
\multicolumn{1}{c}{$\lambda_{\text{var}}$} &
\multicolumn{1}{c}{$\lambda_{\text{inv}}$} &
\multicolumn{1}{c}{$\lambda_{\text{cov}}$} &
\multicolumn{1}{c}{Accuracy} &
\multicolumn{1}{c}{F1 Score} &
\multicolumn{1}{c}{Equalized Odds} &
\multicolumn{1}{c}{Absolute Odds} &
\multicolumn{1}{c}{Statistical Parity} &
\multicolumn{1}{c}{Disparate Impact} \\
\midrule
 &
  0.10 & 0.90 & 0.00 & 0.00 &
   0.8481 $\pm$ 0.0038&
   0.6650 $\pm$ 0.0180&
   0.1018 $\pm$ 0.0270&
   0.0919 $\pm$ 0.0229&
   -0.1943 $\pm$ 0.0285&
   0.3079 $\pm$ 0.0370\\ 
 &
  0.10 & 0.00 & 0.90 & 0.00 &
   0.8150 $\pm$ 0.0053&
   0.4281 $\pm$ 0.0938&
   0.0437 $\pm$ 0.0363&
   0.0342 $\pm$ 0.0330&
   -0.0811 $\pm$ 0.0516&
   0.3199 $\pm$ 0.0383\\ 
 &
  0.10 & 0.00 & 0.00 & 0.90 &
   0.8106 $\pm$ 0.0109&
   0.4396 $\pm$ 0.0764&
   0.3535 $\pm$ 0.1011&
   0.1990 $\pm$ 0.0591&
   0.0282 $\pm$ 0.0291&
   1.4377 $\pm$ 0.6835\\
 &
  0.10 & 0.45 & 0.45 & 0.00 &
   0.8322 $\pm$ 0.0103&
   0.5382 $\pm$ 0.0747&
   0.0608 $\pm$ 0.0254&
   0.0481 $\pm$ 0.0210&
   -0.1112 $\pm$ 0.0332&
   0.3353 $\pm$ 0.0390\\ 
 &
  0.10 & 0.45 & 0.00 & 0.45 &
   0.8275 $\pm$ 0.0079&
   0.5535 $\pm$ 0.0524&
   0.2860 $\pm$ 0.0505&
   0.1597 $\pm$ 0.0294&
   -0.0126 $\pm$ 0.0239&
   0.9651 $\pm$ 0.2578\\ 
\multirow{-6}{*}{\makecell{Adult\\Income}} &
  0.10 & 0.00 & 0.45 & 0.45 &
   0.8137 $\pm$ 0.0048&
   0.4614 $\pm$ 0.0504&
   0.3573 $\pm$ 0.0669&
   0.1975 $\pm$ 0.0380&
   0.0191 $\pm$ 0.0325&
   1.2790 $\pm$ 0.2316\\
\midrule
 &
  0.10 & 0.90 & 0.00 & 0.00 &
   0.6598 $\pm$ 0.0144&
   0.6250 $\pm$ 0.0283&
   0.2932 $\pm$ 0.1011&
   0.2490 $\pm$ 0.0746&
   -0.2834 $\pm$ 0.0733&
   0.6144 $\pm$ 0.0823\\ 
 &
  0.10 & 0.00 & 0.90 & 0.00 &
   0.6571 $\pm$ 0.0121&
   0.6232 $\pm$ 0.0384&
   0.2618 $\pm$ 0.0412&
   0.2101 $\pm$ 0.0266&
   -0.2435 $\pm$ 0.0264&
   0.6530 $\pm$ 0.0551\\
 &
  0.10 & 0.00 & 0.00 & 0.90 &
   0.6475 $\pm$ 0.0172&
   0.6018 $\pm$ 0.0405&
   0.0874 $\pm$ 0.0522&
   0.0606 $\pm$ 0.0427&
   -0.0146 $\pm$ 0.0686&
   1.0010 $\pm$ 0.1556\\
 &
  0.10 & 0.45 & 0.45 & 0.00 &
   0.6683 $\pm$ 0.0103&
   0.6424 $\pm$ 0.0156&
   0.2173 $\pm$ 0.0353&
   0.1809 $\pm$ 0.0239&
   -0.2223 $\pm$ 0.0223&
   0.6694 $\pm$ 0.0392\\ 
 &
  0.10 & 0.45 & 0.00 & 0.45 &
   0.6575 $\pm$ 0.0131&
   0.6147 $\pm$ 0.0280&
   0.1007 $\pm$ 0.0519&
   0.0730 $\pm$ 0.0471&
   -0.0540 $\pm$ 0.0795&
   0.9274 $\pm$ 0.1440\\ 
 \multirow{-6}{*}{\makecell{COMPAS}} &
  0.10 & 0.00 & 0.45 & 0.45 &
   0.6718 $\pm$ 0.0164&
   0.6358 $\pm$ 0.0326&
   0.2047 $\pm$ 0.0448&
   0.1635 $\pm$ 0.0404&
   -0.2018 $\pm$ 0.0454&
   0.7067 $\pm$ 0.0578\\
\midrule
 &
  0.10 & 0.90 & 0.00 & 0.00 &
   0.7140 $\pm$ 0.0253&
   0.8011 $\pm$ 0.0233&
   0.1414 $\pm$ 0.0566&
   0.0951 $\pm$ 0.0412&
   -0.1049 $\pm$ 0.0412&
   0.8646 $\pm$ 0.0511\\ 
 &
  0.10 & 0.00 & 0.90 & 0.00 &
   0.7165 $\pm$ 0.0356&
   0.7917 $\pm$ 0.0319&
   0.1367 $\pm$ 0.0798&
   0.0964 $\pm$ 0.0521&
   -0.0600 $\pm$ 0.1090&
   0.9113 $\pm$ 0.1625\\ 
 &
  0.10 & 0.00 & 0.00 & 0.90 &
   0.7060 $\pm$ 0.0325&
   0.7911 $\pm$ 0.0358&
   0.1497 $\pm$ 0.0863&
   0.1005 $\pm$ 0.0599&
   0.0229 $\pm$ 0.0755&
   1.0225 $\pm$ 0.1128\\
 &
  0.10 & 0.45 & 0.45 & 0.00 &
   0.7485 $\pm$ 0.0436&
   0.8248 $\pm$ 0.0340&
   0.1675 $\pm$ 0.0866&
   0.1135 $\pm$ 0.0546&
   -0.1204 $\pm$ 0.0996&
   0.8469 $\pm$ 0.1271\\ 
 &
  0.10 & 0.45 & 0.00 & 0.45 &
   0.7110 $\pm$ 0.0311&
   0.7975 $\pm$ 0.0231&
   0.1219 $\pm$ 0.0620&
   0.0880 $\pm$ 0.0533&
   0.0175 $\pm$ 0.0635&
   1.0283 $\pm$ 0.0894\\  
\multirow{-6}{*}{\makecell{German\\Credit}} &
  0.10 & 0.00 & 0.45 & 0.45 &
   0.7260 $\pm$ 0.0167&
   0.8091 $\pm$ 0.0145&
   0.1516 $\pm$ 0.0617&
   0.1190 $\pm$ 0.0481&
   -0.0739 $\pm$ 0.1093&
   0.9006 $\pm$ 0.1405\\
\bottomrule
\end{tabular}
}
\end{table}

\end{document}